\documentclass{ieeetmlcn}
\usepackage{cite}
\usepackage{amsmath,amssymb,amsfonts}
\usepackage{algorithmic}
\usepackage{graphicx,color}
\usepackage{textcomp}
\usepackage{xcolor}
\usepackage{hyperref}
\usepackage{gensymb}
\usepackage{multirow}
\hypersetup{hidelinks=true}
\usepackage{algorithm,algorithmic}
\usepackage{array,subcaption}
\usepackage{bbm}

\def\BibTeX{{\rm B\kern-.05em{\sc i\kern-.025em b}\kern-.08em
    T\kern-.1667em\lower.7ex\hbox{E}\kern-.125emX}}
\AtBeginDocument{\definecolor{tmlcncolor}{cmyk}{0.93,0.59,0.15,0.02}\definecolor{NavyBlue}{RGB}{0,86,125}}

\def\OJlogo{\vspace{-4pt}\includegraphics[height=18pt]{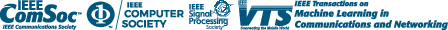}}
\def\seclogo{\vspace{10pt}\includegraphics[height=18pt]{new_logo}}

\def\authorrefmark#1{\ensuremath{^{\textbf{#1}}}}

\begin{document}
\receiveddate{XX Month, XXXX}
\reviseddate{XX Month, XXXX}
\accepteddate{XX Month, XXXX}
\publisheddate{XX Month, XXXX}
\currentdate{XX Month, XXXX}
\doiinfo{TMLCN.2022.1234567}

\markboth{Adaptive Learning Strategies for AoA-Based Outdoor Localization: A Comprehensive Framework}{Bac TN {et al.}}

\title{Adaptive Learning Strategies for AoA-Based Outdoor Localization: A Comprehensive Framework}

\author{Bac Trinh-Nguyen\authorrefmark{1}\authorrefmark{2}, Graduate Student Member, IEEE, Sara Berri\authorrefmark{1}, Member, IEEE,\\ Sin G. Teo\authorrefmark{2}, Tram Truong-Huu\authorrefmark{3}, Senior Member, IEEE\\ and Arsenia Chorti\authorrefmark{1}\authorrefmark{4}, Senior Member, IEEE}
\affil{ETIS UMR 8051, CY Cergy Paris University, ENSEA, CNRS, FR}
\affil{Institute for Infocomm Research, Agency for Science, Technology and Research (A*STAR), SG}
\affil{Singapore Institute of Technology (SIT), SG}
\affil{Barkhausen Institut gGmbH, Dresden, DE}
\corresp{Corresponding author: Bac Trinh-Nguyen (email: trinh-nguyen.bac@ensea.fr).}
\authornote{
B. Trinh-Nguyen has been partially supported by the EC through the Horizon Europe/JU SNS project ROBUST-6G (Grant Agreement no. 101139068), the  Horizon Europe COST Action Project 6G-PHYSEC, the CNRS IPAL Project CONNECTING. S. Berri have been partially supported by the EC through the Horizon Europe/JU SNS project ROBUST-6G (Grant Agreement no. 101139068), the EU HORIZON MSCA-SE TRACE-V2X project (Grant No. 101131204), the ANR-PEPR 5G Future Networks projects. A. Chorti has been partially supported by the EC through the Horizon Europe/JU SNS project ROBUST-6G (Grant Agreement no. 101139068), the  Horizon Europe COST Action Project 6G-PHYSEC, the CNRS IPAL Project CONNECTING, the ENSEA SRV project RETRO, the CYU TalCyb Chair in Cybersecurity and by the French government under the France 2030 ANR program “PEPR Networks of the Future” (ref. ANR-22-PEFT-0008 HISEC and ANR-22-PEFT-0009 FOUNDS). The authors would like to thank S. Wesemann, G. Kaltbeitzel, D. Wiegner, M. Kinzler, S. Merk and S. Woerner from Nokia for realizing the channel measurements and sharing the data.
}

\begin{abstract}

Localization in 5G / B5G and 6G networks is essential for important use cases such as intelligent transportation, smart factories, and smart cities. Although deep learning has enabled improving localization accuracy, depending on the deployment scenario and the effort required  for dataset collection  campaigns on a given infrastructure, the training process for localization models can vary significantly. Furthermore, with respect to feature selection, recent works have demonstrated the robustness of angle-of-arrival (AoA)-based localization.  In view of these two points, we propose an adaptive framework for AoA-based localization that consists of two alternative learning strategies, each suited either for ``large'' or ``small'' training datasets. The proposed framework is evaluated on a real, massive multiple-input multiple-output (mMIMO) orthogonal frequency division multiplexing (OFDM) outdoor channel state information (CSI) dataset. First, we investigate offline learning when large training datasets are available; we propose a hierarchical framework that first distinguishes between line of sight (LoS) and non line of sight (NLoS) regions and then moves to more fine grained localization in the respective region. This approach provides high-performance localization through accumulated batch retraining and an integrated hyperparameter optimization mechanism, achieving $100\%$ accuracy in distinguishing LoS and NLoS regions, $99.82\%$ accuracy for predefined trajectories classification in the LoS region and approximately $98\%$ accuracy for those in the NLoS region. Second, when only a small training dataset is available, an online learning framework is proposed, using incremental tree-based and ensemble-based models for handling streaming data and continuously updating mode, as well as an online few-shot learning model for rapidly initializing new classes from a limited labeled support set. The aggregated Mondrian Forest (AMF) achieves an accuracy of approximately $94\%$ for the classification of trajectories in both LoS and NLoS regions, with very low forgetting rates ranging from $0.0248$ to $0.0427$. These results showcase that robust localization in outdoor wireless environments is achievable with low-latency, and, demonstrate that high accuracies can be achieved incrementaly during network operation by exploiting online learning, alleviating the need for large dataset collection campaigns.

\end{abstract}

\begin{IEEEkeywords}
6G, wireless localization, channel state information, angle of arrival, machine learning, continual learning, online learning, few-shot learning, generative models.
\end{IEEEkeywords}


\maketitle
\section{INTRODUCTION}
\label{sec:introduction}
\IEEEPARstart{L}{ocalization} in 5G / B5G and 6G networks is an essential capability due to its critical role in various applications such as smart industry, autonomous agents, and intelligent transportation systems. In modern multiple input multiple output (MIMO) systems, channel state information (CSI) is measured between transmitting and receiving devices and can be used for localization tasks~\cite{hsieh2019deep,chen2017confi}. Recent works have discussed the robustness of CSI features against noise and adversarial attacks and the potentially important role of robust localization in enhancing security and trustworthiness~\cite{10336902, 11456828} of wireless systems. Among different trusted features for localization, earlier works relied on time of arrival (ToA) and time difference of arrival (TDoA). 

Recently, the works in ~\cite{pham2023machine,  Murali24  ,  Pham26 , skaperas26 } have shown that angular-based positioning techniques such as angle-of-arrival (AoA) and angle-of-departure (AoD) not only provide a scalable and cost-effective approach with significant potential for future  applications~\cite{fischer2025systematic}, but also, importantly, cannot be forged in digital array MIMO systems and can serve as \emph{robust} physical layer features for localization and location-based authentication. 

However in real-world wireless systems, the fluctuations of estimated AoA values due to noise, multipath propagation and interference are generally non negligible. Although this complicates their direct use for localization, AoA estimates still provide discriminative patterns that can be exploited as powerful features in machine learning (ML)-based localization methods. Combining AoA features with supervised ML techniques has been shown to achieve strong localization performance in both indoor and outdoor scenarios~\cite{yao2025high,liu2025deep}, reconfirming that ML and deep learning (DL) have facilitated enhanced localization accuracy. 

In the following, we present an overview of several localization learning approaches, including offline learning, online learning, and few-shot learning, highlighting current literature gaps and how our proposed framework aims at closing them. We then also briefly review our early results on the proposed AoA-based localization offline-learning framework, followed by the presentation of our main contributions on a comprehensive framework of apative learning for AoA-based outdoor localization.

\subsection{OFFLINE LEARNING-BASED LOCALIZATION}
In recent years, ML techniques have been widely adopted for radio frequency (RF)-based localization to overcome the limitations of traditional approaches, including sensitivity to environmental dynamics, hardware imperfections, and synchronization constraints~\cite{intro_burghal2020comprehensive}. For example, a method based on $k$-nearest neighbors (KNN) was shown in~\cite{intro_sobehy2020csi} to achieve high localization accuracy using CSI in indoor environments. The work of Yang \textit{et al.}~\cite{intro_yangsurvey} further provided a comprehensive survey of ML-driven indoor localization, highlighting the use of supervised learning techniques such as support vector machines (SVMs), KNN and neural networks within offline fingerprinting frameworks.

Beyond indoor scenarios, MobIntel~\cite{bao2021mobintel} demonstrated the effectiveness of ML-based methods for passive outdoor localization using received signal strength indicator (RSSI) measurements. In addition, multipath-assisted CSI fingerprinting approaches have been proposed in~\cite{chen2018multipath}, achieving meter-level accuracy in outdoor localization settings. These works validated the strong potential of ML-based techniques in diverse RF localization environments. However, these approaches do not distinguish between LoS and NLoS propagation environments, which have well-known and distinct impacts on achievable localization accuracy due to their differing multipath structures, instead relying on black-box learning to predict location.

\subsection{ONLINE LEARNING-BASED LOCALIZATION}
Online learning, also known as continual or incremental learning, aims to update models sequentially with incoming data streams while mitigating catastrophic forgetting and avoiding complete retraining from scratch. As a result, it can reduce computational cost and resource consumption compared to batch retraining approaches. Due to mobility, fading, and environmental changes in wireless and CSI-based systems, continual learning has been applied to some tasks, such as channel estimation and action recognition~\cite{akrout2023continual, zhang2023csi, zhang2025carec, mohsin2025continual}.

Elastic weight consolidation (EWC)~\cite{kirkpatrick2017overcoming} is one of regularization-based methods that constrains updates on parameters identified as important for previous tasks. Memory-aware synapses (MAS)~\cite{aljundi2018memory} and replay-based methods such as iCaRL~\cite{rebuffi2017icarl} retain a subset of past samples and add them during the training process on new tasks to mitigate forgetting. These methods typically rely on backpropagation, stored samples, and epoch-based retraining on historical data, which may limit their applicability in streaming and resource-constrained wireless scenarios.

In contrast, online and streaming learning focuses on single-pass, incremental model updates when samples arrive sequentially, while storage and retraining are treated as optional supporting mechanisms rather than core requirements. Within this paradigm, tree-based incremental models such as the hoeffding tree (HT)~\cite{hulten2001mining} leverage statistical guarantees to grow decision trees from streaming data, while adaptive variants like the hoeffding adaptive tree (HAT)~\cite{bifet2009adaptive} incorporate drift detection to handle non-stationary environments. Ensemble methods improve robustness and accuracy by combining multiple incremental models, such as adaptive random forests (ARF)~\cite{gomes2017adaptive}, streaming random patches (SRP)~\cite{gomes2019streaming}, and aggregated mondrian forests (AMF)~\cite{mourtada2021amf}, which exploit model diversity and drift awareness. These ensemble and streaming tree-based approaches are well suited to wireless localization scenarios where CSI distributions evolve over time in dynamic environments. 

In this work, we utilize incremental models implemented through the River library~\cite{montiel2021river} to enable continual adaptation under streaming data conditions in an AoA-based online learning framework. We rely on these built-in configurations without introducing an external drift detection module. However, the use of these online learning approaches in AoA-based localization is still limited, particularly in terms of balancing adaptation capability, computational efficiency, and reliability of localization decisions under varying propagation conditions.

\subsection{FEW-SHOT LEARNING}
Few-shot learning aims to enable models to perform new tasks using only a small number of labeled examples. Meta-learning, or learning to learn, provides an effective solution for few-shot learning by training models across multiple tasks to acquire transferable knowledge that allows rapid adaptation to unseen tasks with limited data~\cite{gharoun2023}. Existing meta-learning approaches for few-shot learning can be categorized into metric-based methods such as prototypical networks (ProtoNet)~\cite{snell2017prototypical}, memory-based methods, and learning-based methods such as model-agnostic meta-learning (MAML)~\cite{finn2017model}. 

In the context of wireless sensing and localization, several recent works have explored meta-learning techniques for fast adaptation across environments. Few-shot learning was investigated for Wi-Fi-based indoor positioning using CNN and meta-learning techniques, demonstrating that meta-learning can achieve competitive performance under limited data conditions~\cite{xie2024few}. MetaLoc~\cite{gao2023metaloc} introduced the MAML-based fingerprinting localization framework that learned meta-parameters from multiple well-calibrated domains. Cui \textit{et al.}~\cite{cui2025profi} proposed ProFi-Net, a prototype-based few-shot Wi-Fi gesture recognition method that extended Prototypical Networks with feature-level attention and gradually increased noise-based query augmentation to improve robustness and generalization. 

Few-shot learning for AoA estimation using prototypical networks~\cite{mashaal2025lightweight} demonstrated strong performance under limited data and domain shifts, highlighting the potential of few-shot approaches for adaptive localization systems. Nevertheless, although previous works have explored the use of few-shot learning techniques, such as ProtoNet, in wireless communication tasks, their application to angular-based localization remains limited.

\subsection{DATA AUGMENTATION}
Due to the high cost of collecting labeled wireless data, data scarcity remains a major challenge in training robust localization models. Generative models have emerged as effective solutions for synthesizing additional training data, with generative adversarial networks (GANs)~\cite{goodfellow2020generative} and variational autoencoders (VAEs)~\cite{pinheiro2021variational} being among the most widely used approaches to generating realistic synthetic samples. To further enhance class-conditional generation, conditional variants, such as conditional GANs (CGANs)~\cite{mirza2014conditional} and conditional VAEs (CVAEs)~\cite{sohn2015learning} incorporate label information during the generation process. These generative models have been applied to enrich the CSI datasets and improve generalization under limited data conditions~\cite{wang2025generative, feng2025survey}. Previous studies have shown that generative data augmentation can enhance model robustness to environmental variations and measurement noise, particularly in offline training scenarios.

\subsection{Proposed Adaptive Localization Framework}

Although prior studies have explored offline learning, online learning, and few-shot learning for wireless localization, important gaps still remain. Existing offline approaches typically assume relatively fixed environments and sufficient labeled training data, which may limit their applicability in dynamic deployment conditions. Online learning approaches offer the ability to incrementally update models as new data arrive, but their potential for AoA-based localization has not been sufficiently investigated. Similarly, few-shot learning has shown promise for rapid adaptation with limited labeled samples, yet it is often studied in isolation rather than as part of a broader adaptive localization strategy. Moreover, previous studies considered offline learning, continual learning, few-shot learning as isolated solutions. Therefore, there is still a need for an adaptive localization framework that can support both offline and online learning paradigms under different scenarios while leveraging AoA-based representations for practical real-world deployment.

In our previous work~\cite{trinhhigh}, we extracted AoA features from CSI measurements using multiple signal classification (MUSIC)~\cite{music_schmidt1986multiple} and estimation of signal parameters using rotational invariance techniques (ESPRIT)~\cite{roy1989esprit}. We employed a hierarchical classification strategy that first discriminates between LoS and NLoS regions, followed by two region-specific classifiers to identify predefined trajectories within each region. In~\cite{trinhhigh}, we systematically assessed six baseline machine learning models: $k$-nearest neighbors (KNN), random forest (RF), logistic regression (LR), gradient boosting machine (GBM), extreme gradient boosting (XGBoost), light gradient boosting machine (LightGBM), as well as the stacking ensemble model~\cite{wolpert1992stacked}. 
Although the baseline localization results were already strong, in the current work we achieve further performance gains through hyperparameter optimization of classifiers, where accuracy is used as the optimization objective. 

Despite the near perfect performance, offline supervised learning requires a sufficiently large and representative labeled dataset and typically requires full retraining when the channel characteristics, user trajectories, or LoS / NLoS conditions vary over time, non-stationary data distributions. The models trained in an offline manner are unable to fully capture all these variations and therefore require periodic retraining or when performance monitoring modules detect degradation below a predefined threshold. During this process, all collected data must be stored and labeled prior to retraining, which increases system latency and resource consumption.

Even more importantly, offline learning relies on the availability of large labeled datasets, which might be impractical in many radio access network (RAN) deployment scenarios and limits its adaptability to dynamic environments and newly observed classes. To address these shortcoming, we investigate two online learning approaches, including continual tree- and ensemble tree-based methods, as well as continual few-shot learning, to address the buring issue of limited data availability. We adopt a metric-based few-shot learning strategy inspired by ProtoNet for scenarios in which only a small number of labeled samples are available at a time. The proposed approach enables rapid initialization of newly observed trajectories or regions from limited labeled data, which is important for practical wireless localization in limited-label streaming environments.

While prior studies typically treat offline learning, continual learning, and few-shot learning as isolated approaches, this work proposes a unified framework for AoA-based localization that systematically leverages these paradigms across different operational phases of the network. The proposed framework aligns naturally with the deployment lifecycle of future 6G RAN systems. 
In the initial phase, where labeled data are scarce, online learning combined with few-shot techniques enables rapid model initialization and early adaptation. As additional data become available at the gNodeBs and are aggregated through cloud or virtual RAN pooling, continual learning facilitates progressive model refinement under streaming conditions. In later stages, once sufficient labeled data have been collected, offline training can be used to derive a more accurate and stable model. Overall, this results in a scalable, adaptive, and robust localization solution suited for dynamic wireless environments.



The contributions of this paper are summarized as follows:
\begin{itemize}
    \item We design an offline AoA-based localization framework based on high-accuracy hierarchical classifiers. In the first stage, a binary classifier distinguishes LoS from NLoS regions with $100\%$ accuracy. In the second stage, region-specific classifiers achieve $99.82\%$ accuracy in the LoS region and approximately $98\%$ accuracy in the NLoS region after hyperparameter tuning.
    \item Next, we propose an online learning framework that leverages incremental tree-based and ensemble-based classifiers to process streaming data, along with continual few-shot learning to quickly initialize new classes with limited labeled samples, thereby enabling the adaptive operation under dynamic conditions. The results demonstrate that AMF is the most effective online learner for AoA-based localization, achieving about $94\%$ accuracy in both LoS and NLoS regions, with very low forgetting rates between $0.0248$ and $0.0427$.
    \item In addition, we propose a conditional variational autoencoder (CVAE) for data augmentation in the offline setting to simulate potential environmental variations and evaluate the robustness of our proposed methods.
\end{itemize}

The remainder of the paper is organized as follows. Section~\ref{sec:framework} presents our proposed adaptive localization framework. Section~\ref{sec:experiments} presents the experimental setup and analysis of experimental results. Finally, Section~\ref{sec:conclusion} concludes the paper and discusses future directions.

\section{PROPOSED AOA-BASED LOCALIZATION FRAMEWORK}
\label{sec:framework}

\begin{figure*}
    \centering
    \includegraphics[width=\linewidth]{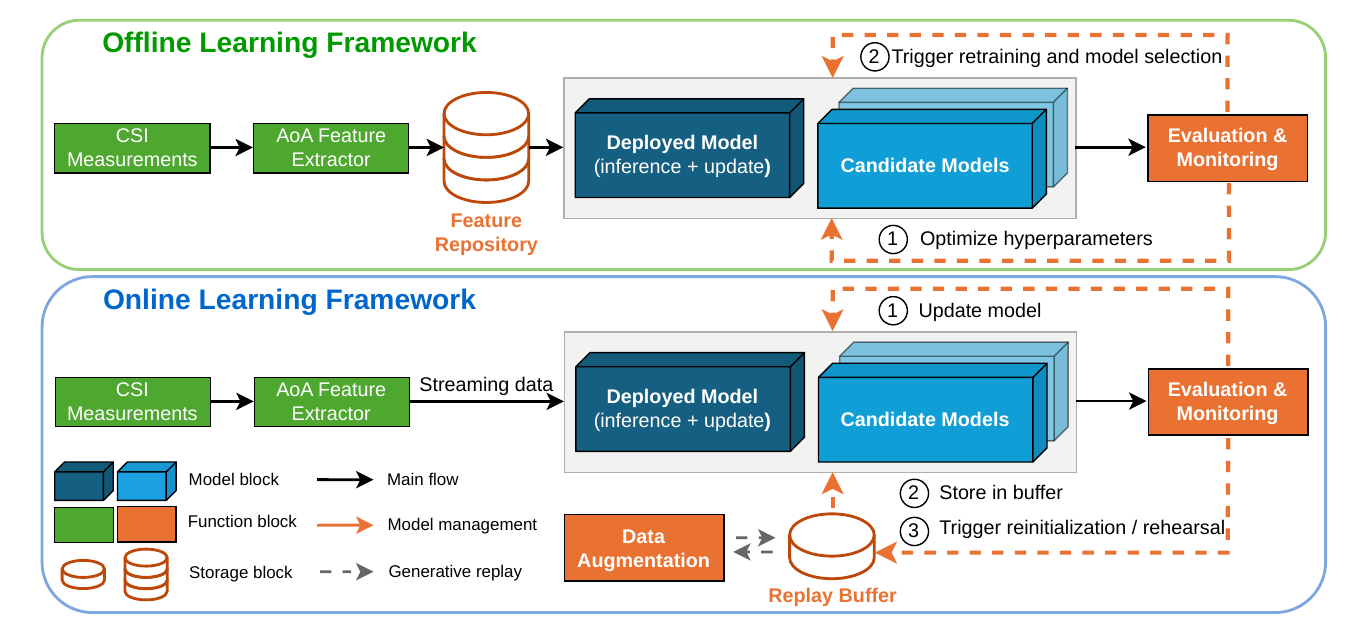}
    \caption{Overview of AoA-based localization framework. Offline learning performs batch training and retraining with model selection, while online learning handles streaming data and updates models incrementally.}
    \label{fig:two_lane_overview}
\end{figure*}



To enable robust localization in wireless communications, we propose an adaptive AoA-based localization framework that integrates both offline and online learning strategies depending on the deployment scenario while meeting the requirements for low-latency and real-time applications. Fig.~\ref{fig:two_lane_overview} illustrates the proposed adaptive localization framework. In the offline learning block, the system is initialized through offline training using AoA features extracted from CSI measurements. When deployed, trained models provide a strong performance baseline with high reliability. 

To ensure stable deployment, we introduce a parallel model evaluation and replacement mechanism along with the hyperparameter optimization module. The deployed model remains unaffected and operates in inference mode. New  AoA features extracted from CSI measurements and labeled samples are accumulated and used to train candidate models. Candidate models that demonstrate better performance than the current model are promoted to deployment. This design ensures reliable localization over time.

In scenarios where latency and resource are critical, the online learning approach operates using continual learning and few-shot learning. This is particularly important when only limited data is available at deployment time, such as when a new base station is installed, where large-scale data collection is impractical. The adaptive models can be pre-trained with a small amount of labeled data to facilitate smooth model initialization, enable gentle parameter adjustment, and help the model generalize faster and more accurately during the inference phase. During the inference phase, the model parameters are updated incrementally as new data samples arrive. To mitigate performance degradation caused by concept drift or catastrophic forgetting, the scheme incorporates a small replay buffer and a monitoring mechanism. The online learning workflow is as follows:
\begin{enumerate}
    \item The model is updated incrementally as new data arrive.
    \item Newly acquired data are stored in a replay buffer.
    \item If data drift is detected or the model exhibits catastrophic forgetting, the evaluation module triggers a rehearsal or selective model reinitialization.
\end{enumerate}

Since we investigate two approaches within the online learning framework, the details of continual learning and the few-shot learning approach are described in Section~\ref{sec:framework} Part~\ref{sec:framework:online_learning_framework}.

\subsection{OFFLINE LEARNING LOCALIZATION FRAMEWORK}
\label{sec:framework:offline_learning}
\begin{figure}
    \centering
    \includegraphics[width=0.98\linewidth]{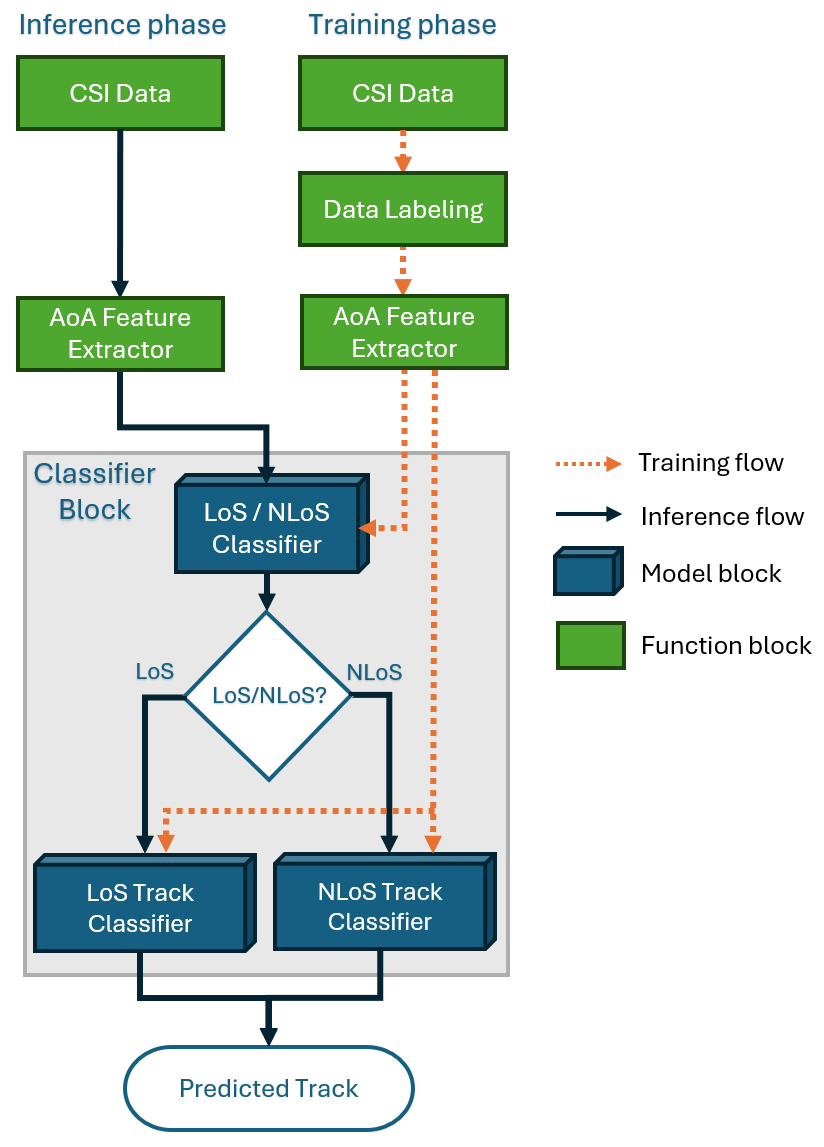}
    \caption{Architecture of hierarchical two-stage classifiers (Offline Learning Framework).}
    \label{fig:hierarchical_framework}
\end{figure}

The hierarchical two-stage classification framework is illustrated in Fig.~\ref{fig:hierarchical_framework}. It consists of two main stages. In the first stage, a binary classifier discriminates between LoS and NLoS regions. In the second stage, two region-specific multiclass classifiers distinguish fine-grained predefined trajectories (tracks) within each region. These classifiers are trained and evaluated prior to deployment in two main phases: the training phase and the inference phase.

\textbf{Training phase.} The training phase aims to develop and select models for deployment and operation. CSI measurements are preprocessed and labeled by regions and track identifier (track ID). Next, MUSIC and ESPRIT algorithms are applied to estimate AoAs and construct feature vectors, which are then fed into the classifier block to train a hierarchical classifier including a binary LoS / NLoS classifier, followed by two region-specific multi-class classifiers: one for LoS tracks and one for NLoS tracks. During training, we evaluate multiple ML algorithms, including LR, KNN, RF, GBM, XGBoost, LightGBM, and a stacking ensemble model (combining the top $n$ best performing ML models, which $n \in {2,...,6}$. Finally, the best-performing and most robust models are selected for deployment.

\textbf{Inference phase.} Incoming CSI samples are processed in real-time to extract AoA features. This will first be classified by the LoS / NLoS classifier to determine to which region the incoming data belong to. The corresponding region-specific track classifier is then used to identify the exact track. If a sample is classified as LoS, the LoS track classifier predicts the track ID. Otherwise, it is forwarded to the NLoS track classifier to predict the track ID.

\textbf{Hyperparameter Optimization.} To improve classification performance without altering the overall localization framework, a monitoring and evaluation module is added. This module continuously assesses the performance of the deployed model and triggers accumulative batch retraining of candidate models when performance degradation is detected or when sufficient new data have been collected. During this process, model hyperparameters are optimized using the accumulated dataset. A trigger mechanism is used to promote the deployment of a candidate model when the performance of the deployed model degrades and the candidate model achieves superior performance. This process serves as a backup solution in the event of an unexpected model failure.

There are several methods for hyperparameter tuning in ML, including manual tuning where hyperparameters are selected heuristically, exhaustive search strategies such as grid search, random search, and Bayesian optimization~\cite{owen2022hyperparameter}. In this work, we investigate Bayesian optimization and perform automated hyperparameter tuning using Optuna~\cite{akiba2019optuna} for the candidate classifiers introduced in Section~\ref{sec:framework:offline_learning}. Instead of brute-force combinations of values, Bayesian optimization formulates the tuning process as a sequential optimization problem. It leverages information from previous evaluations to guide the search toward promising regions of the hyperparameter space and determines subsequent trials accordingly.

The final goal is to identify well-performing configurations that can be used consistently in the deployment phase. For each classifier, we define an objective function that samples a hyperparameter configuration from a predefined search space as described in Table~\ref{tbl:searchspace} and evaluates generalization performance using 5-fold cross-validation on the training set. The optimization process aims to maximize the mean cross-validation accuracy across folds.

\subsection{ONLINE LEARNING LOCALIZATION FRAMEWORK}
\label{sec:framework:online_learning_framework}

In this subsection, we present a data augmentation mechanism in Part~\ref{sec:framework:online_learning_framework:data_augmentation}, which employs generative models to support model training and evaluation under data scarcity conditions. We introduce our online learning framework through several approaches. First, we describe a baseline online learning strategy based on periodic batch retraining with conventional ML models. Next, we present incremental learning in streaming settings using tree-based classifiers and ensemble models (Part~\ref{sec:framework:online_learning_framework:online_learning}). Finally, we investigate few-shot learning with prototypical networks as a representative approach to enable rapid adaptation when only a small number of labeled samples are available for new tracks (Part~\ref{sec:framework:online_learning_framework:few_shot_learning}).

\subsubsection{Data Augmentation using Generative Models}\label{sec:framework:online_learning_framework:data_augmentation}

\begin{figure}
    \centering
    \includegraphics[width=\linewidth]{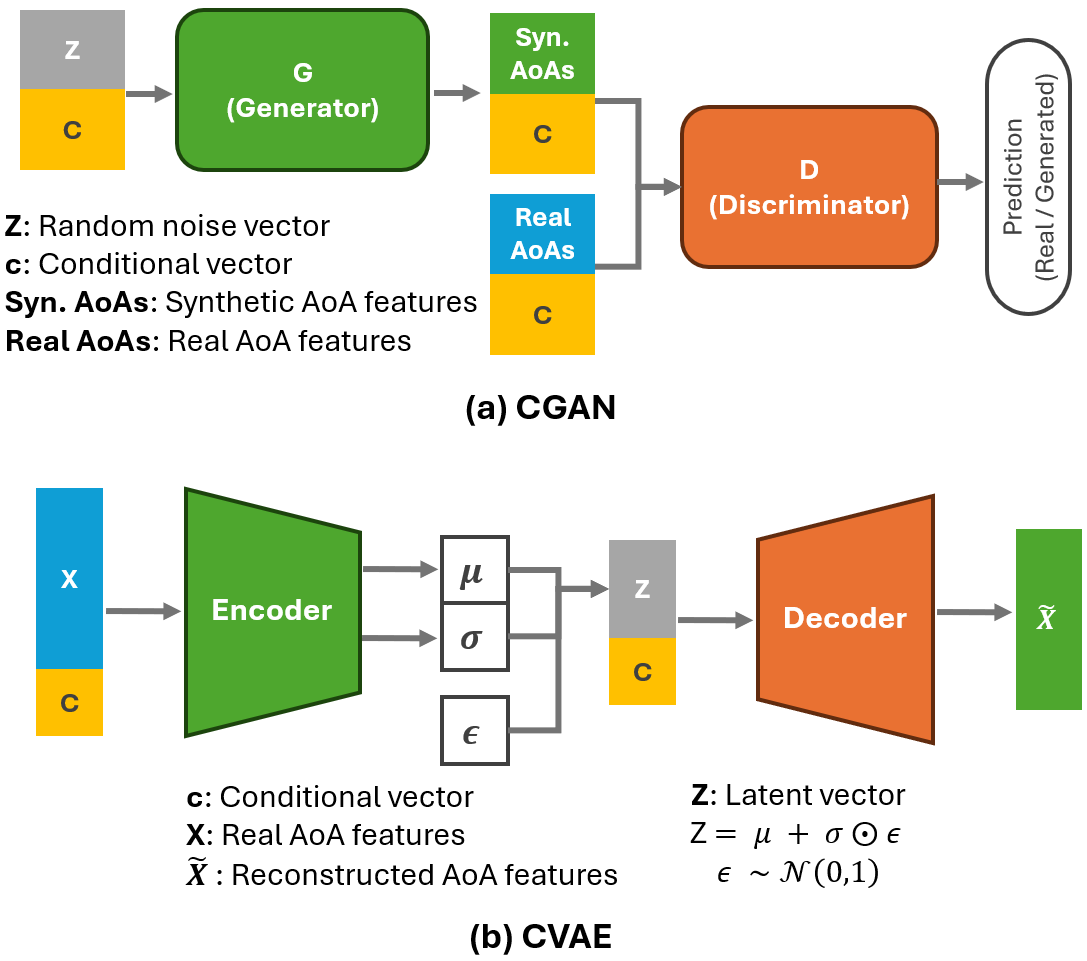}
    \caption{Architectures of the Conditional Generative Adversarial Network (CGAN) and the Conditional Variational Autoencoder (CVAE) used for feature-level AoA data augmentation. Both models operate on AoA feature vectors conditioned on class labels to generate / reconstruct synthetic AoA samples.}
    \label{fig:cgan_cvae}
\end{figure}

Even with strong fine-tuned classifiers, the current offline approach may still degrade in real-world deployments due to environmental dynamics, delayed or unavailable labels, and the computational cost of retraining models from scratch to adapt with distribution shifts. To address these limitations, we investigate a data augmentation module based on two conditional generative models for feature-level augmentation: conditional generative adversarial network (CGAN) and conditional variational autoencoder (CVAE). This module generates synthetic samples from the original data with two main objectives:
\begin{itemize}
    \item To mimic realistic variations in the AoA feature distribution supporting model evaluation and improving robustness under previously unseen conditions,
    \item To support upsampling of replay buffer during rehearsal or reinitialization processes.
\end{itemize}
Their architectures are presented in Fig.~\ref{fig:cgan_cvae}. In both cases, generation is conditioned on a label vector $c$ that encodes the scenario information, including the propagation region (LoS / NLoS) and the corresponding predefined track IDs. Table~\ref{tbl:CGAN_CVAE_architecture} summarizes the architectures of these two generative models evaluated for feature-level augmentation. Both models are implemented as multi-layer perceptrons (MLPs). Specifically, CGAN consists of an MLP-based generator and an MLP-based discriminator, while CVAE consists of an MLP-based encoder and an MLP-based decoder.

\textbf{Conditional Generative Adversarial Network (CGAN).}
The conditional GAN comprises a generator $G$ and a discriminator / critic $D$. Let $x$ denote a real sample drawn from the data distribution, conditioned on the label vector $c$ (encoding LoS / NLoS and track ID), the generator synthesizes AoA feature vectors $\tilde{x}=G(z,c)$ from a noise input $z$, aiming to generate class-consistent samples that resemble real AoA features. The critic $D(\cdot,c)$ assigns higher scores to real samples than to generated ones. Unlike standard GANs, the Wasserstein GAN (WGAN)~\cite{arjovsky2017wasserstein} replaces the cross-entropy with the Wasserstein distance, providing smoother gradients and more stable training. In addition, we apply a gradient penalty (WGAN-GP)~\cite{gulrajani2017improved} to enforce the 1-Lipschitz constraint. The generator $G$ is trained to minimize loss 
\begin{equation}
\mathcal{L}_G = -\mathbb{E}[D(\tilde{x},c)],    
\end{equation}
and the critic is trained by minimizing 
\begin{equation}
\mathcal{L}_D = -\mathbb{E}[D(x,c)] + \mathbb{E}[D(\tilde{x},c)] + \lambda_{\mathrm{gp}}\mathrm{GP},    
\end{equation}
where $\lambda_{\mathrm{gp}}$ controls the contribution of the gradient penalty (GP) term 
\begin{equation}
    \mathrm{GP} = \mathbb{E}_{\hat{x}} \left[ \left( ||\nabla_{\hat{x}} D(\hat{x}, c)||_2 - 1 \right)^2 \right]
\end{equation}
where $\hat{x}$ denotes an interpolated sample between a real sample $x$ and a generated sample $\tilde{x}$, used to evaluate the gradient penalty.

\textbf{Conditional Variational Autoencoder (CVAE).}
CVAE is structured by two main components, the encoder and the decoder. Given an AoA feature vector $x$ and its conditional label $c$, the encoder $q_{\phi}(z|x,c)$ maps the concatenated input $[x, c]$ to the parameters of a Gaussian latent distribution $\mu_{\phi}(x,c)$ and $\log \sigma^2_{\phi}(x,c)$. A latent vector is then obtained via the reparameterization trick~\cite{kingma2015variational}:
\begin{equation}
    z=\mu + \sigma \odot \epsilon, \quad \epsilon\sim\mathcal{N}(\mathbf{0},\mathbf{I}),      
\end{equation}
where $\odot$ denotes element-wise multiplication.

The decoder $p_{\theta}(x\mid z,c)$ takes the latent vector $z$ and conditional label $c$ to reconstruct $\tilde{x}$. Training minimizes the negative evidence lower bound (ELBO) which consists of a reconstruction term, implemented as the mean squared error (MSE) and a Kullback-Leibler (KL) divergence regularization term:
\begin{equation}
\begin{aligned}
\mathcal{L}_{\mathrm{VAE}}
=&\; \mathcal{L}_{\mathrm{recon}} + \mathcal{L}_{\mathrm{KL}} \\
=&\; \mathbb{E}_{q_{\phi}(z\mid x, c)}\!\left[\lVert x-\tilde{x}\rVert_2^2\right] + \\
&\;\beta\, D_{\mathrm{KL}}\!\left(q_{\phi}(z\mid x,c)\,\|\,p(z)\right),
\end{aligned}
\end{equation}
where $D_{\mathrm{KL}}(\cdot|\cdot)$ is the Kullback–Leibler divergence and $\beta$ controls the strength of the KL regularizer.

The CVAE and CGAN models are evaluated using a realistic outdoor dataset, with detailed experimental results reported in Section~\ref{sec:experiments} Part~\ref{sec:experiments:data_augmentation}. As illustrated in Fig.~\ref{fig:two_lane_overview}, the trained CVAE-based data augmentation module is connected to the replay buffer in the practical implementation. Based on these results, CVAE produces more stable training behavior and high quality synthetic AoA features in our setting. Therefore, we apply CVAE as the primary data augmentation method for the remainder of the online learning strategy. This design enables data upsampling even when only a small set of samples is available, thus facilitating quick model reinitialization and supporting evaluation after rehearsal using the replay buffer.

\subsubsection{Continual Learning}
\label{sec:framework:online_learning_framework:online_learning}

In real-world deployments, CSI / AoA data arrive sequentially as a data stream. Labels may be missing, delayed, or even only available sparsely, while environmental dynamics can induce distribution shifts over time. An ideal localization framework should maintain stable performance under non-stationary conditions while minimizing the computational cost of repeated full retraining.

\textbf{Batch Retraining with Conventional ML.} We extend the conventional ML pipeline introduced in Section~\ref{sec:framework} Part~\ref{sec:framework:offline_learning} to a streaming-like setting by periodically retraining models on accumulated data batches (e.g., every day or every week). We consider two batch retraining approaches:
\begin{itemize}
    \item \textbf{Buffer retraining:} model is retrained using only the most recent data window (current buffer).
    \item \textbf{Cumulative retraining:} model is retrained using all samples observed so far.
\end{itemize}

Although the cumulative training approach provides strong performance, it has clear practical limitations:
\begin{itemize}
    \item Batch requirement and latency: the system must wait until enough samples are collected before retraining, leading to delay adaptation.
    \item Lack of streaming updates: conventional offline learners are not designed for incremental updates and typically require full retraining.
    \item Computational and storage cost: since data needs to be stored, this increases retraining time and memory requirements.
\end{itemize}

These limitations motivate the need for a continual / incremental method that supports online optimization of model parameters without full retraining, adapts to evolving data distributions, and reduces computation and storage. In the following, we evaluate a set of incremental models including tree-based incremental models and ensemble methods.

\textbf{Continual / Incremental Learning.} To support deployment in dynamic environments, we adopt a continual learning approach. In contrast to batch training, online learning updates the model incrementally as new samples arrive, enabling adaptation to potential distribution shifts due to environmental changes and user's mobility. These observations motivate us to move beyond batch retraining toward a learning paradigm that can continuously update as data arrive. Therefore, we formulate AoA-based localization as a data-stream problem and employ continual classifiers that perform lightweight incremental updates. In this work, we evaluate several continual learning models, including:
\begin{itemize}
    \item \textbf{Hoeffding Tree Classifier (HT):} Incremental decision tree for data streams that grows using the Hoeffding bound to decide splits from a limited number of observations, as defined:
    \begin{equation}
        \epsilon = \sqrt{\frac{R^2 \ln(1/\delta)}{2n}}, 
    \end{equation} 
    where $n$ denotes the number of samples observed at a node, $R$ is the range of the chosen split evaluation metric, and $\delta$ is the confidence parameter that limits the probability of selecting a suboptimal split.
    \item \textbf{Hoeffding Adaptive Tree Classifier (HAT):} Drift-adaptive extension of Hoeffding trees that monitors predictive performance of branches using a drift detector and can replace subtrees when degradation is detected.
    \item \textbf{Adaptive Random Forest Classifier (ARF):} Streaming ensemble of incremental trees designed for non-stationary streams and uses per-tree drift detection to selectively reset or replace underperforming trees.
    \item \textbf{Streaming Random Patches Classifier (SRP):} Online ensemble method that generalizes bagging / random-subspaces for streams. The default base learner is a Hoeffding Tree, though other incremental estimators can be used.
    \item \textbf{Aggregated Mondrian Forest (AMF):} Online forest based on Mondrian-tree partitions, supporting single-pass learning and anytime prediction. Each node maintains a regularized label distribution, and predictions can be aggregated along the path to the leaf. The forest outputs the average of class probabilities across trees.
    \item \textbf{Gaussian Naive Bayes (GNB):} Lightweight incremental probabilistic baseline that maintains a Gaussian per class and feature. Predictions are computed via the summed per-feature log-likelihoods under the Naive Bayes independence assumption, making it fast and memory-efficient.
\end{itemize}
 

\textbf{Streaming flow and online update procedure.} The continual learning framework operates under a streaming setting in real-time deployment, including an initial warm-up phase and a subsequent online inference and update stage.

\begin{itemize}
    \item Streaming flow: we consider that a data stream $x_t$ denotes the AoA feature vector extracted at time $t$ and $y_t$ is the associated label.
    \item Warm-up training (model initialization): we initialize each classifier using the first 10\% of the dataset that assumes a realistic setting in which only a small labeled dataset ready at the beginning.
    \item Online phase (predict-then-train streaming task): For the 90\% of dataset, we will implement the predict-then-train strategy for simulating online learning, each data stream $x_t$ will arrive and the classifier tries to predict and learn from this new data:
    \begin{enumerate}
        \item Predict: $\hat{y}_t = f_{t-1}(\mathbf{x}_t)$ using the current model $f_{t-1}$.
        \item Evaluate: update the streaming metrics using ($\hat{y}_t$ and $y_t$).
        \item Learn: update the model with the new labeled sample $f_t$ $\xleftarrow{}$ update($f_{t-1}$, $x_t$, $y_t$).
    \end{enumerate}
	This phase mirrors online deployment, where decisions must be made immediately, and the model improves as labeled feedback becomes available.
\end{itemize}

\subsubsection{Few-shot Learning}
\label{sec:framework:online_learning_framework:few_shot_learning}

While continual learning supports efficient adaptation to streaming data when the set of trajectories (tracks / classes) is fixed and labels are available regularly, it does not fully address settings in which samples from previously unseen trajectories appear over time. In such cases, continual classifiers may require substantial labeled data to reliably update their decision boundaries, resulting in slow adaptation and potentially unstable updates. To address this gap, we therefore investigate an online few-shot meta-learning strategy based on prototypical networks (ProtoNet). ProtoNet is a metric-based method that can form a new class from $K$ labeled samples by constructing class prototypes in the learned embedding space and supports rapid initialization of unseen classes.

We first train and evaluate ProtoNet under the standard episodic few-shot meta-learning setup, and then extend it to an online setting by updating class prototypes incrementally as new labeled samples arrive.

\textbf{Support Set and Query Set.} In meta-learning, the dataset $D$ is split into two sets: meta-train ($D_{meta\_train}$) and meta-test ($D_{meta\_test}$), corresponding to train and test sets in supervised ML. During the training and testing processes, the meta-train and the meta-test are again divided into train and test sets. Thus, to avoid confusion with the conventional train / test terminology used within each episode (task), we use the terms support set and query set. Finally, the datasets are partitioned as follows:
\begin{align}
D_{meta\_train} &= \{ D^{support}_{meta\_train}, D^{query}_{meta\_train} \}, \\
D_{meta\_test} &= \{ D^{support}_{meta\_test}, D^{query}_{meta\_test} \}.
\end{align}
where $D_{meta\_train}$ is used to train the meta-learner model and this model is evaluated on episodes sampled from $D_{meta\_test}$.

\textbf{N-way K-shot problem.} The training process in meta-learning proceeds over episodes (tasks). In each episode $\mathcal{T}$, a support set $S$ and a query set $Q$ are constructed. The support set follows the $N$-way $K$-shot setting, which means that it consists of $N$ classes with $K$ labeled samples per class. The query set contains $M$ additional samples for each of the same $N$ classes. Formally,
\begin{align}
S &= \left\{(x_{n,k},\, y_n)\ \middle|\ n\in\{1,\dots,N\},\ k\in\{1,\dots,K\}\right\}, \\
Q &= \left\{(x'_{n,m},\, y_n)\ \middle|\ n\in\{1,\dots,N\},\ m\in\{1,\dots,M\}\right\}.
\end{align}

\textbf{Episodic training.}
During each meta-train iteration, we sample an episode $\mathcal{T}_i=(S_i,Q_i)$ from $D_{meta\_train}$. Given an embedding network $f_{\theta}(\cdot)$, the learner is adapted using only the support set $S_i$ and is evaluated on the query set $Q_i$. Few-shot meta-learning helps model learn to adapt quickly for new tasks with a very few labeled samples, therefore, the learning strategy is different from traditional way that is, instead of learning to classify directly, the model learns how to learn from small datasets.


\begin{figure}
    \centering
    \includegraphics[width=\linewidth]{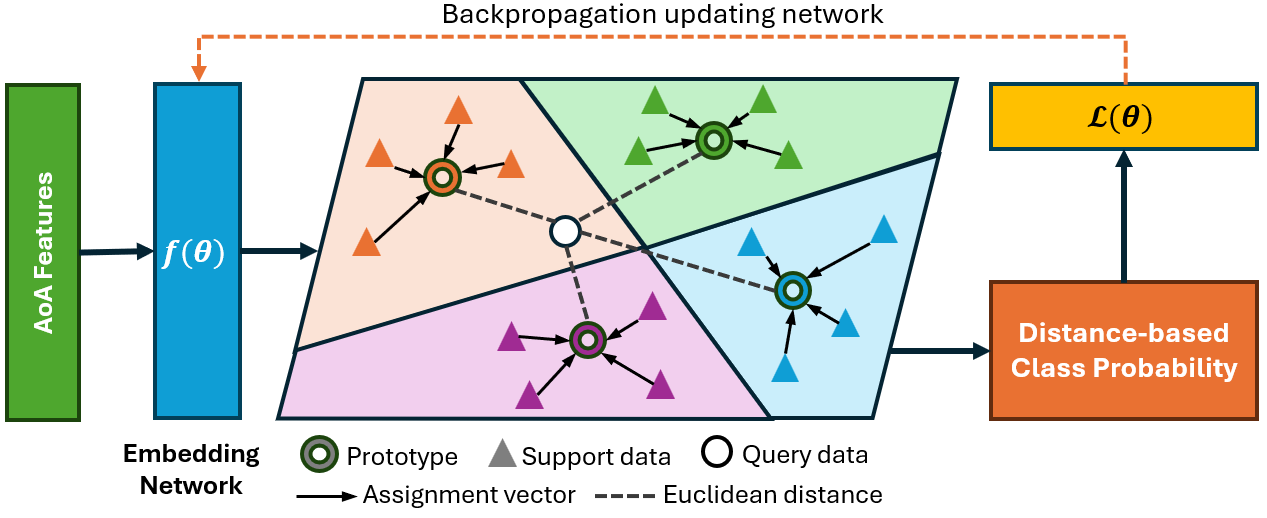}
    \caption{Architecture of prototypical networks (ProtoNet) for AoA-based classification.}
    \label{fig:protonet}
    \vspace{-2.5ex}
\end{figure}

\textbf{Prototypical Networks (ProtoNet).} Prototypical networks (ProtoNet) are a widely used approach for few-shot classification. The key idea of ProtoNet, as illustrated in Fig.~\ref{fig:protonet}, is to represent each class by a prototype, defined as the mean of its support examples in a learned embedding space. This prototype represents the center of each class, enabling rapid class initialization from $K$ labeled examples. ProtoNet uses an embedding network $f_{\theta}(\cdot)$ to map an input $x$ into values in the mapping space: 
\begin{equation}
    z = f_{\theta}(x) \in R^d.
\end{equation}
For each class $c \in \{1,...,N\}$, ProtoNet computes a prototype as the mean embedding of its support set $S_c$:
\begin{equation}
   p_c = \frac{1}{|S_c|} \sum_{(x_i, y_i) \in S_c} f_\theta(x_i).
\end{equation}

Given a new query sample $x$, ProtoNet computes the squared Euclidean distance between the query embedding and the class prototype. Class probabilities are obtained by applying a softmax over negative distances:
\begin{equation}
p_\theta(y=c\mid x)=\frac{\exp\left(-||f_\theta(x)-p_c||_2^2\right)}{\sum_{c'=1}^{N}\exp\left(-||f_\theta(x)-p_{c'}||_2^2\right)}.
\end{equation}
The embedding parameters $\theta$ are learned by minimizing the average negative log-likelihood $\mathcal{L} = -\log p_{theta}(y = k \mid x)$ of the true class $c$ using stochastic gradient descent. 



\textbf{Continual Few-shot Learning via ProtoNet.} Motivated by the fact that class prototypes can be updated incrementally, we apply a ProtoNet-based solution to handle the practical setting where only a small number of labeled examples per class may be available at any time. ProtoNet handles new observed classes differently from incremental tree-based classifiers. Rather than expanding the classifier structure by creating new branches or removing weak branches. ProtoNet performs classification on an embedding space where each class is represented by a prototype. When labeled samples from a new class arrive, a small labeled support set is passed through the embedding network / encoder to obtain their embedding vectors. The corresponding class prototype is computed as the mean embedding of these support samples. This prototype is then added to the prototype set, enabling continual few-shot adaptation without retraining from scratch. For each new incoming query sample, the encoder maps it into the embedding space and computes its distances to all existing prototypes. The query is assigned to the class of the closest prototype.

In the following, we report the experimental setup and results for the proposed AoA-based localization approach. We first present the performance of offline machine learning models before and after improvement through hyperparameter tuning. We then compare generative models for data augmentation. Finally, we analyze the results of online learning approaches with continual tree-based and ensemble models and the few-shot learning approach.

\section{EXPERIMENTS}
\label{sec:experiments}

\subsection{DATASET DESCRIPTION}
In this work, we evaluate our proposed approaches on a real world 64-antenna massive MIMO 50-subcarrier OFDM outdoor dataset at FR1 (2.18 GHz) collected at the Nokia campus, Stuttgart, Germany~\cite{nokia_dataset}. The dataset includes uplink CSI measurements from a massive MIMO digital antenna array, arranged in 4 rows $\times$ 16 columns of single-polarized patch antennas. The antenna array is installed on the roof of a 20 m-high building with a 10 degrees downtilt. The horizontal spacing between adjacent antennas is $\lambda/2$, and the vertical spacing is $\lambda$, at a central frequency of 2.18 GHz. 

\begin{figure*}[ht]
    \centering
    \includegraphics[width=\linewidth]{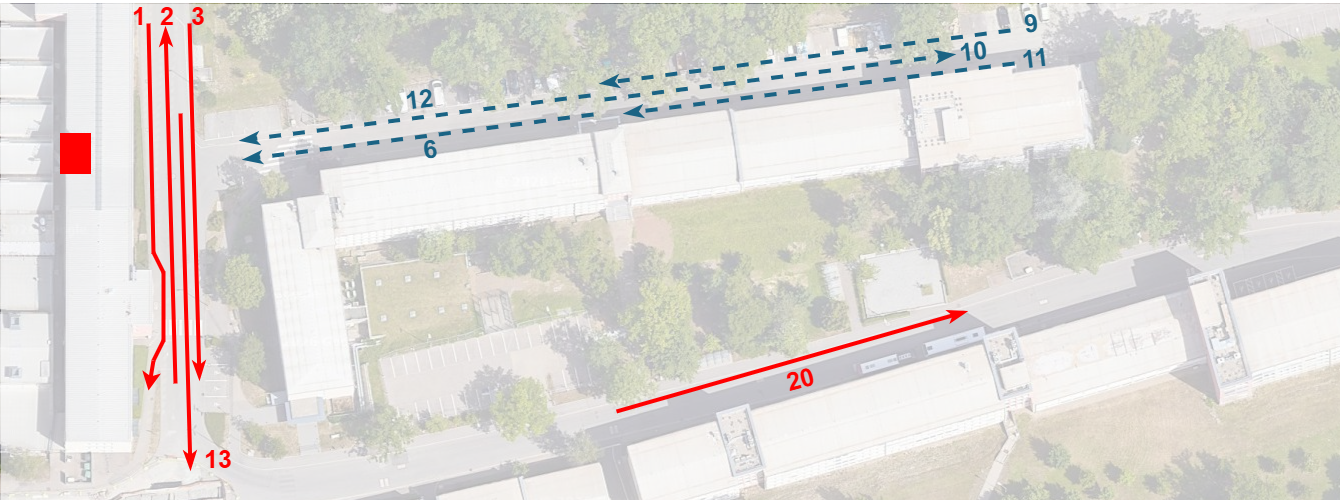}
    \caption{Nokia campus in Stuttgart, Germany. The red rectangle denotes the mMIMO antenna array mounted on top of a building, while the lines with arrows represent the trajectories (tracks) and their respective directions. Red solid lines indicate NLoS tracks, while blue dashed lines represent LoS tracks.}
    \label{fig:nokia_map}
\end{figure*}

50 OFDM subcarriers spanning a 10 MHz bandwidth are recorded per antenna, pilot bursts are transmitted every 0.5 ms. The receiver is a monopole antenna mounted on a cart at an approximate height of 2 meters. The cart follows predefined trajectories (tracks) at an average speed of 3-5 km/h, yielding a spatial sampling interval of 0.5 mm. The size of the dataset for each individual track is reported in Table~\ref{tbl:track_shape}. Due to obstacles such as buildings and trees, the measurement area comprises both LoS and NLoS regions. In this work, we consider five LoS tracks (tracks 6, 9, 10, 11, and 12) and five NLoS tracks (tracks 1, 2, 3, 13, and 20), as shown on the campus map in Fig.~\ref{fig:nokia_map}.

\subsection{FEATURE EXTRACTION}
\label{sec:experiments:feature_extraction}

Let $H \in \mathbb{C}^{N \times M \times T}$ denote the CSI tensor, where $N=64$ is the total number of antennas arranged in 4 rows and 16 columns, $M=50$ is the number of subcarriers and $T$ is the number of consecutive measurement locations along each track, sampled every 0.5 mm. We use the CSI dataset described above to extract AoA vectors across different antenna-array rows and OFDM subcarriers. 
To do so, we apply the MUSIC and ESPRIT algorithms to estimate the azimuth AoA for each data segment. Since the estimation accuracy of both methods depends on the input length, we study its impact on classification by considering window lengths $W \in {500, 1000, 2000}$ consecutive measurements for AoA extraction. This corresponds to CSI spans of ${0.25, 0.5, 1}$ meters for each AoA estimate. To further enlarge the AoA dataset, we slide the window $W$ along each track with a 50$\%$ overlap (shift\_ratio), which ensures smoother transitions between adjacent frames and increases the number of samples. 

To obtain a feature vector from a data segment (window) with $W$ samples along each track, we first estimate four AoAs independently from the four rows of the uniform linear array. We then exploit the 50 OFDM subcarriers to compute 50 AoA estimates per row, resulting in a $4 \times 50 = 200$-dimensional AoA feature vector. As a result, we construct three AoA dataset configurations, summarized in Table~\ref{tbl:track_shape}.

\begin{table}[t]
\small
\centering
\caption{Performance of MUSIC and ESPRIT in AoA estimation across track 11.}
\label{tbl:track_details}
\begin{tabular}{|c|cc|cc|}
\hline
\multirow{2}{*}{\bf Algorithm} & \multicolumn{2}{c|}{\textbf{AoA estimation ($\degree$)}} & \multicolumn{2}{c|}{\textbf{Processing time (s)}}        \\ \cline{2-5} 
                  & \multicolumn{1}{c|}{\textbf{Mean}} & \textbf{Std Dev} & \multicolumn{1}{c|}{\textbf{Total}} & \textbf{Mean} \\ \hline
\textbf{MUSIC}    & \multicolumn{1}{c|}{-35.17}        & 1.71           & \multicolumn{1}{c|}{7.1672}         & 0.0300                    \\ \hline
\textbf{ESPRIT}   & \multicolumn{1}{c|}{-35.30}          & 1.63             & \multicolumn{1}{c|}{0.2777}         & 0.0012                  \\ \hline
\end{tabular}
\vspace{-1.5ex}
\end{table}

\begin{figure}[t]
    \centering
    \includegraphics[width=\linewidth]{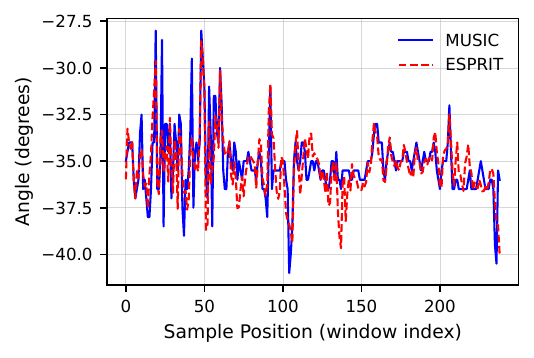}
    \caption{Example of AoA estimation for track 11, using antenna row $0$ at subcarrier $10$, with a window size of $1000$.}
    \label{fig:example_aoa_estimation}
    \vspace{-2.5ex}
\end{figure}

Fig.~\ref{fig:example_aoa_estimation} illustrates an AoA estimation example for track 11 using a window length $W=1000$, at antenna row 0 and subcarrier 10. The mean AoA estimates produced by MUSIC and ESPRIT are almost identical ($-35.17\degree$ and $-35.30\degree$, respectively), with comparable standard deviations of about $1.71\degree$ and $1.63\degree$. The computational costs are reported in Table~\ref{tbl:track_details}, MUSIC requires 7.1672 seconds of processing time which is approximately 26 times longer than ESPRIT (0.2777 seconds). On average, this corresponds to 30 ms per estimate for MUSIC versus 1.2 ms for ESPRIT. 

\begin{table}[!t]
\footnotesize
\centering
\caption{Data shapes of the original CSI and estimated AoAs (MUSIC) for LoS tracks (6, 9, 10, 11, 12) and NLoS tracks (1, 2, 3, 13, 20). The dataset generated using ESPRIT follows the same format.}
\label{tbl:track_shape}
\begin{tabular}{|c|c|ccc|}
\hline
\multirow{2}{*}{\textbf{Tracks}} & \multirow{2}{*}{\textbf{CSI data shape}} & \multicolumn{3}{c|}{\textbf{AoA data shape by window size}}                            \\ \cline{3-5} 
                                 &                                          & \multicolumn{1}{c|}{\textbf{500}} & \multicolumn{1}{c|}{\textbf{1000}} & \textbf{2000} \\ \hline
1                                & (60, 50, 120000)                         & \multicolumn{1}{c|}{(479, 200)}   & \multicolumn{1}{c|}{(239, 200)}    & (119, 200)    \\ \hline
2                                & (60, 50, 120000)                         & \multicolumn{1}{c|}{(479, 200)}   & \multicolumn{1}{c|}{(239, 200)}    & (119, 200)    \\ \hline
3                                & (60, 50, 122000)                         & \multicolumn{1}{c|}{(487, 200)}   & \multicolumn{1}{c|}{(243, 200)}    & (121, 200)    \\ \hline
13                               & (60, 50, 122000)                         & \multicolumn{1}{c|}{(487, 200)}   & \multicolumn{1}{c|}{(243, 200)}    & (121, 200)    \\ \hline
20                               & (60, 50, 118000)                         & \multicolumn{1}{c|}{(471, 200)}   & \multicolumn{1}{c|}{(235, 200)}    & (117, 200)    \\ \hline
6                                & (60, 50, 96000)                          & \multicolumn{1}{c|}{(383, 200)}   & \multicolumn{1}{c|}{(191, 200)}    & (95, 200)     \\ \hline
9                                & (60, 50, 120000)                         & \multicolumn{1}{c|}{(479, 200)}   & \multicolumn{1}{c|}{(239, 200)}    & (119, 200)    \\ \hline
10                               & (60, 50, 120000)                         & \multicolumn{1}{c|}{(479, 200)}   & \multicolumn{1}{c|}{(239, 200)}    & (119, 200)    \\ \hline
11                               & (60, 50, 120000)                         & \multicolumn{1}{c|}{(479, 200)}   & \multicolumn{1}{c|}{(239, 200)}    & (119, 200)    \\ \hline
12                               & (60, 50, 92000)                          & \multicolumn{1}{c|}{(367, 200)}   & \multicolumn{1}{c|}{(183, 200)}    & (91, 200)     \\ \hline
\end{tabular}
\end{table}

\subsection{OFFLINE LEARNING PERFORMANCE}
\subsubsection{Performance of LoS / NLoS Classifier (Stage 1)}

We evaluate the performance of six models: LR, KNN, RF, GBM, LightGBM, XGBoost, while progressively increasing the training-set size from 5\% to 80\% in steps of 5\%. Table~\ref{tbl:first_stage} reports the first-stage LoS / NLoS classification results using the AoA features estimated by MUSIC. All evaluated models achieved perfect performance with an accuracy of 1. These results indicate that, given sufficient training data, the first-stage classifier preserves perfect accuracy and does not degrade subsequent LoS and NLoS track classifiers in the second stage. Even with only 5$\%$ of the training data, all models exceeded the accuracy of 97$\%$, highlighting the strong discriminative power of the AoA-based features.

Notably, KNN and RF reached 100\% accuracy with just 5\% training data (about 50 samples) for all AoA window sizes tested, $W \in \{500,1000,2000\}$. Both models required substantially less training time than the remaining methods, highlighting their efficiency. However, KNN incurs a noticeably higher inference latency than LR, RF, or GBM. For real-time scenarios where inference speed is critical, LR and GBM offer an attractive trade-off, with inference times of 0.24-0.26 ms and 0.78-0.82 ms, respectively.

\begin{table}[t]
\small
\centering
\caption{Model performance (accuracy, F1-score, and ROC AUC), training time in seconds (T (s)), and inference time in milliseconds (I (ms)) across different window sizes (W).}
\label{tbl:first_stage}
\begin{tabular}{|l|c|c|c|r|r|}
\hline
\textbf{Model (W)} & \textbf{Acc.} & \textbf{F1} & \textbf{AUC} & \textbf{T (s)} & \textbf{I (ms)} \\
\hline
LR (2000)       & 1    & 1    & 1   & 0.029  & 0.239 \\
\hline
LR (1000)       & 1    & 1    & 1   & 0.123  & 0.238 \\
\hline
LR (500)        & 1    & 1    & 1   & 0.315  & 0.264 \\
\hline
KNN (2000)      & 1    & 1    & 1   & 0.002  & 72.657 \\
\hline
KNN (1000)      & 1    & 1    & 1   & 0.002  & 77.847 \\
\hline
KNN (500)       & 1    & 1    & 1   & 0.004  & 83.951 \\
\hline
RF (2000)       & 1    & 1    & 1   & 0.377  & 12.485 \\
\hline
RF (1000)       & 1    & 1    & 1   & 0.569  & 11.896 \\
\hline
RF (500)        & 1    & 1    & 1   & 1.003  & 11.929 \\
\hline
GBM (2000)      & 0.99 & 0.99 & 1   & 3.439  & 0.794 \\
\hline
GBM (1000)      & 0.99 & 0.99 & 1   & 7.804  & 0.817 \\
\hline
GBM (500)       & 0.99 & 0.99 & 1   & 17.426 & 0.824 \\
\hline
LightGBM (2000) & 1    & 1    & 1   & 19.498 & 11.020 \\
\hline
LightGBM (1000) & 1    & 1    & 1   & 27.386 & 11.485 \\
\hline
LightGBM (500)  & 1    & 1    & 1   & 40.610 & 11.850 \\
\hline
XGBoost (2000)  & 1    & 1    & 1   & 26.473 & 53.174 \\
\hline
XGBoost (1000)  & 1    & 1    & 1   & 28.684 & 53.324 \\
\hline
XGBoost (500)   & 1    & 1    & 1   & 28.540 & 52.469 \\
\hline
\end{tabular}
\vspace{-2ex}
\end{table}

\subsubsection{Performance of Track Classifiers (Stage 2)}

The results and the corresponding training and inference times for the ML models are summarized in Table~\ref{tbl:train_inference_times}. 

Overall, tree ensemble methods (RF, GBM, XGBoost, and LightGBM) achieve the strongest results for both MUSIC- and ESPRIT-based AoA features. The stacking ensembles (constructed by combining the top $n$ base classifiers, with $n \in \{2,\ldots,6\}$ consistently achieve the best classification performance, particularly when paired with ESPRIT. In most cases, ESPRIT provides a higher accuracy than MUSIC as the AoA estimator. Stacking the top-6 model has the highest results (98\% with MUSIC, 95.4$\%$ with ESPRIT). While lightweight models such as LR and KNN offer very fast inference, they yield lower accuracy, reaching 71.1$\%$ and 76.6$\%$ with MUSIC, respectively. Among individual classifiers, RF provides the most attractive trade-off: it achieves high accuracy (94.3$\%$ with MUSIC and 92.6$\%$ with ESPRIT) while maintaining a low training time (0.52–0.76 s) and a low inference time (approximately 11 ms).

\begin{table}[t]
\footnotesize
\centering
\caption{The comparison of accuracy, training time, and inference time across ML models for MUSIC and ESPRIT (window size = 2000), sorted by accuracy.}
\begin{tabular}{|lccc|}
\hline
\multicolumn{1}{|c|}{\textbf{Models}} & \multicolumn{1}{c|}{\textbf{Accuracy (\%)}} & \multicolumn{1}{c|}{\textbf{Training (s)}} & \textbf{Inference (s)} \\ \hline
\multicolumn{4}{|c|}{\textbf{MUSIC}}                                                                                                         \\ \hline
\multicolumn{1}{|l|}{Stacking Top-6}  & \multicolumn{1}{c|}{97.98}             & \multicolumn{1}{r|}{1023.9869}         & 0.0657             \\ \hline
\multicolumn{1}{|l|}{RF}              & \multicolumn{1}{c|}{94.30}             & \multicolumn{1}{r|}{0.5239}            & 0.0110             \\ \hline
\multicolumn{1}{|l|}{GBM}             & \multicolumn{1}{c|}{89.46}             & \multicolumn{1}{r|}{10.6841}           & 0.0018             \\ \hline
\multicolumn{1}{|l|}{XGBoost}         & \multicolumn{1}{c|}{89.10}             & \multicolumn{1}{r|}{117.3329}          & 0.0212             \\ \hline
\multicolumn{1}{|l|}{LightGBM}        & \multicolumn{1}{c|}{85.98}             & \multicolumn{1}{r|}{103.7215}          & 0.0063             \\ \hline
\multicolumn{1}{|l|}{KNN}             & \multicolumn{1}{c|}{76.62}             & \multicolumn{1}{r|}{0.0020}            & 0.0383             \\ \hline
\multicolumn{1}{|l|}{LR}              & \multicolumn{1}{c|}{71.09}             & \multicolumn{1}{r|}{0.6774}            & 0.0003             \\ \hline
\multicolumn{4}{|c|}{\textbf{ESPRIT}}                                                                                                        \\ \hline
\multicolumn{1}{|l|}{Stacking Top-6}  & \multicolumn{1}{c|}{95.38}             & \multicolumn{1}{r|}{1260.2022}         & 0.0693             \\ \hline
\multicolumn{1}{|l|}{RF}              & \multicolumn{1}{c|}{92.61}             & \multicolumn{1}{r|}{0.7599}            & 0.0108             \\ \hline
\multicolumn{1}{|l|}{LR}              & \multicolumn{1}{c|}{86.73}             & \multicolumn{1}{r|}{0.7763}            & 0.0002             \\ \hline
\multicolumn{1}{|l|}{KNN}             & \multicolumn{1}{c|}{85.06}             & \multicolumn{1}{r|}{0.0020}            & 0.0374             \\ \hline
\multicolumn{1}{|l|}{XGBoost}         & \multicolumn{1}{c|}{84.69}             & \multicolumn{1}{r|}{120.0175}          & 0.0211             \\ \hline
\multicolumn{1}{|l|}{GBM}             & \multicolumn{1}{c|}{83.41}             & \multicolumn{1}{r|}{28.4534}           & 0.0017             \\ \hline
\multicolumn{1}{|l|}{LightGBM}        & \multicolumn{1}{c|}{82.32}             & \multicolumn{1}{r|}{137.4263}          & 0.0044             \\ \hline
\end{tabular}
\label{tbl:train_inference_times}
\vspace{-1ex}
\end{table}

\subsubsection{Hyperparameter Optimization}

The hyperparameter optimization module is integrated into the offline learning framework and automatically fine-tunes the models using the available dataset from the feature repository. This step helps identify optimal hyperparameters, therefore improving model robustness and stability. Since the Stage-1 LoS / NLoS classifier achieves perfect accuracy on our dataset, we do not further optimize this classifier. Instead, we focus on the Stage-2 region-specific multi-class classifiers. In the Appendix, Table~\ref{tbl:searchspace} presents the tuned parameters and their search space. Table~\ref{tbl:hyperparameters} reports the best hyperparameter configurations identified by the Optuna framework~\cite{akiba2019optuna} for each classifier in the LoS and NLoS regions and for each AoA estimator (MUSIC and ESPRIT).

For each model, Optuna runs 500 trials and selects the configuration that maximizes the mean accuracy of 5-fold cross-validation on the training data. For each Optuna-tuned configuration, we retrain each classifier on the full training set and evaluate it once on the held-out test set. We compare the accuracy of the baseline region-specific track classifier proposed in~\cite{trinhhigh} and the Optuna hyperparameter optimization for both MUSIC and ESPRIT algorithms in Table~\ref{tbl:optuna_compare}. The results show that fine-tuning improves the performance of most models in both LoS and NLoS regions, with larger gains in NLoS with boosted ensembles (GBM / XGBoost / LightGBM). The best fine-tuned performance is achieved by 99.82\% in LoS (RF, MUSIC) and 97.99\% in NLoS (GBM, MUSIC), indicating that carefully tuned ensemble methods can improve model performance even under NLoS propagation.

While the proposed offline localization framework combined with hyperparameter optimization achieves high accuracy under offline conditions, it does not fully capture the challenges of real-world deployments. In practice, outdoor propagation environments are highly dynamic, which can induce distribution shifts between training and operational data. Moreover, labels are not readily available for incoming CSI measurements, making the labeling process costly and time-consuming. These factors can degrade performance after deployment and motivate the need for models that can continuously update from streaming measurements, learn effectively from a few labeled samples, and remain stable without catastrophic forgetting.

\begin{table}[t]
\caption{Stage 2 region-specific track classification accuracy (\%) using AoA features (2000 snapshots). Baseline (settings of \cite{trinhhigh}) vs Optuna-tuned configurations. "-" indicates configurations not evaluated in the baseline \cite{trinhhigh}.}
\label{tbl:optuna_compare}
\begin{tabular}{|l|c|cc|cc|}
\hline
\multirow{2}{*}{\textbf{Region}} & \multirow{2}{*}{\textbf{Model}} & \multicolumn{2}{c|}{\textbf{MUSIC}}                      & \multicolumn{2}{c|}{\textbf{ESPRIT}}                     \\ \cline{3-6} 
                                 &                                 & \multicolumn{1}{c|}{\textbf{Baseline}} & \textbf{Optuna} & \multicolumn{1}{c|}{\textbf{Baseline}} & \textbf{Optuna} \\ \hline
\multirow{7}{*}{LoS}             & LR                              & \multicolumn{1}{c|}{71.1}              & 74.4            & \multicolumn{1}{c|}{86.7}              & 88.4            \\ \cline{2-6} 
                                 & KNN                             & \multicolumn{1}{c|}{76.6}              & 77.72           & \multicolumn{1}{c|}{85.1}              & 92.07           \\ \cline{2-6} 
                                 & SVM                             & \multicolumn{1}{c|}{-}                 & 98.53           & \multicolumn{1}{c|}{-}                 & \textbf{99.81}  \\ \cline{2-6} 
                                 & RF                              & \multicolumn{1}{c|}{94.3}              & \textbf{99.82}  & \multicolumn{1}{c|}{92.6}              & 97.05           \\ \cline{2-6} 
                                 & GBM                             & \multicolumn{1}{c|}{89.5}              & 99.63           & \multicolumn{1}{c|}{83.4}              & 94.85           \\ \cline{2-6} 
                                 & XGBoost                         & \multicolumn{1}{c|}{89.1}              & 97.79           & \multicolumn{1}{c|}{84.7}              & 93.74           \\ \cline{2-6} 
                                 & LightGBM                        & \multicolumn{1}{c|}{86}                & 97.98           & \multicolumn{1}{c|}{82.3}              & 92.27           \\ \hline
\multirow{7}{*}{NLoS}            & LR                              & \multicolumn{1}{c|}{50.1}              & 54.11           & \multicolumn{1}{c|}{60.4}              & 59.8            \\ \cline{2-6} 
                                 & KNN                             & \multicolumn{1}{c|}{72.4}              & 72.2            & \multicolumn{1}{c|}{90.3}              & 91.46           \\ \cline{2-6} 
                                 & SVM                             & \multicolumn{1}{c|}{-}                 & 79.73           & \multicolumn{1}{c|}{-}                 & 95.65           \\ \cline{2-6} 
                                 & RF                              & \multicolumn{1}{c|}{87.1}              & 95.81           & \multicolumn{1}{c|}{87.4}              & 95.81           \\ \cline{2-6} 
                                 & GBM                             & \multicolumn{1}{c|}{71.2}              & \textbf{97.99}  & \multicolumn{1}{c|}{68.3}              & \textbf{96.64}  \\ \cline{2-6} 
                                 & XGBoost                         & \multicolumn{1}{c|}{75}                & 88.62           & \multicolumn{1}{c|}{73.7}              & 89.11           \\ \cline{2-6} 
                                 & LightGBM                        & \multicolumn{1}{c|}{74.2}              & 88.27           & \multicolumn{1}{c|}{75.4}              & 89.45           \\ \hline
\end{tabular}
\end{table}

\subsection{CONTINUAL / INCREMENTAL LEARNING}
\label{sec:experiments:incremental_learning}

\subsubsection{Data Augmentation}
\label{sec:experiments:data_augmentation}

In this section, we evaluate the effectiveness of conditional generative models CGAN and CVAE using a real outdoor CSI dataset. We aim to assess training stability, the quality of the AoA features generated, and their impact on localization performance under limited data conditions.

\begin{figure}
    \centering
    \includegraphics[width=\linewidth]{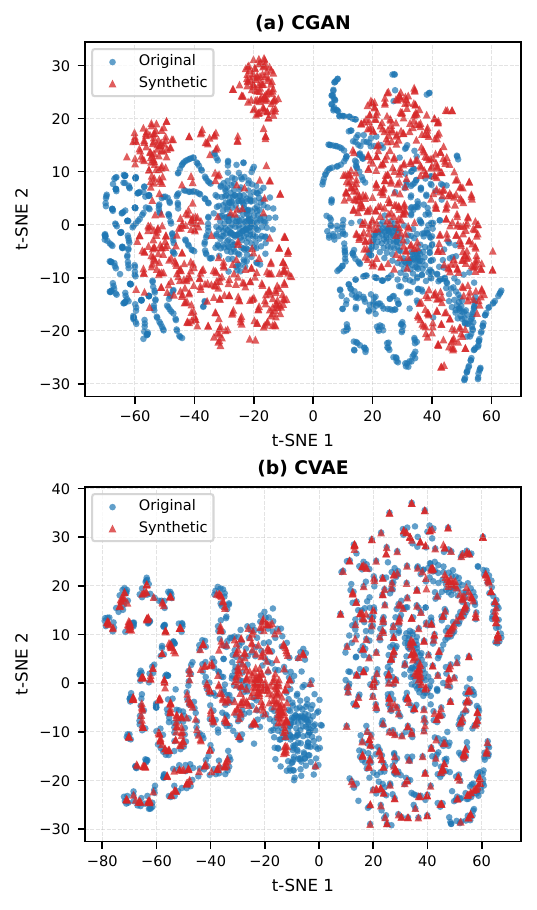}
    \caption{Compares original and synthetic samples distributions between CGAN and CVAE.}
    \label{fig:CGAN_CVAE_compare}
    \vspace{-2.5ex}
\end{figure}

We train CGAN and CVAE models on 10 tracks across the LoS and NLoS regions, using 543 and 597 AoA feature vectors, respectively. The AoA features are extracted using MUSIC and ESPRIT with a window length of $W=2000$ and 50\% overlap. The dataset is split into training and testing sets with an 80 / 20 ratio. During training, the generative models are evaluated at each epoch, and the best-performance models are selected on the basis of validation performance. The t-SNE visualization comparing the distributions of real AoA features estimated by MUSIC and synthetic samples generated by CGAN and CVAE is shown in Fig.~\ref{fig:CGAN_CVAE_compare}. Fig.~\ref{fig:CGAN_CVAE_compare}(a) shows that CGAN-generated samples form a nearby but partially separated cluster, indicating a distribution mismatch between generated and real features. In contrast, CVAE-generated samples in Fig.~\ref{fig:CGAN_CVAE_compare}(b) largely overlap with the real data, confirming that CVAE captures the dominant structure of the feature space. The architecture of CGAN and CVAE is presented in Table~\ref{tbl:CGAN_CVAE_architecture} in the Appendix.

\begin{figure}[t]
    \centering
    \includegraphics[width=\linewidth]{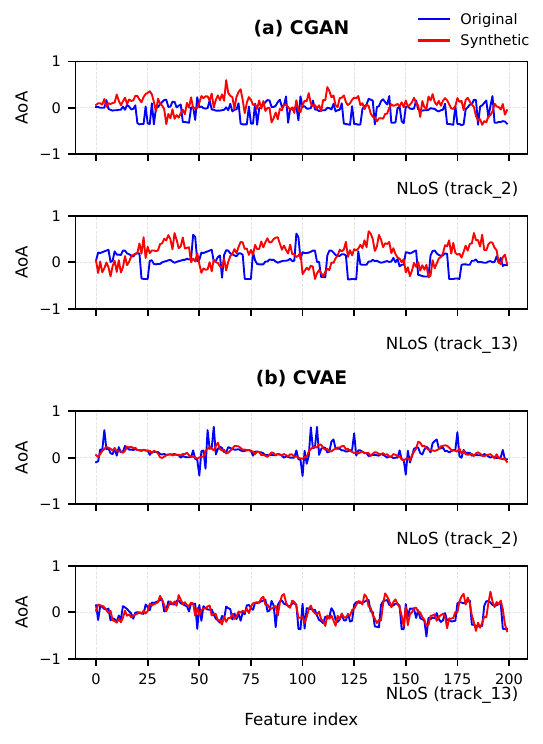}
    \caption{Compares original and synthetic AoA feature patterns between CGAN and CVAE.}
    \label{fig:CGAN_CVAE_patterns}
    \vspace{-2.5ex}
\end{figure}

 Moreover, Fig.~\ref{fig:CGAN_CVAE_patterns} compares the original and synthetic AoA features for several representative track samples. The results show that CVAE reconstructions better preserve the main feature patterns, whereas CGAN outputs have higher variance and more noise artifacts, suggesting that CVAE captures the dominant structure of the feature space.
 
\begin{figure*}[t]
    \centering
    \includegraphics[width=\linewidth]{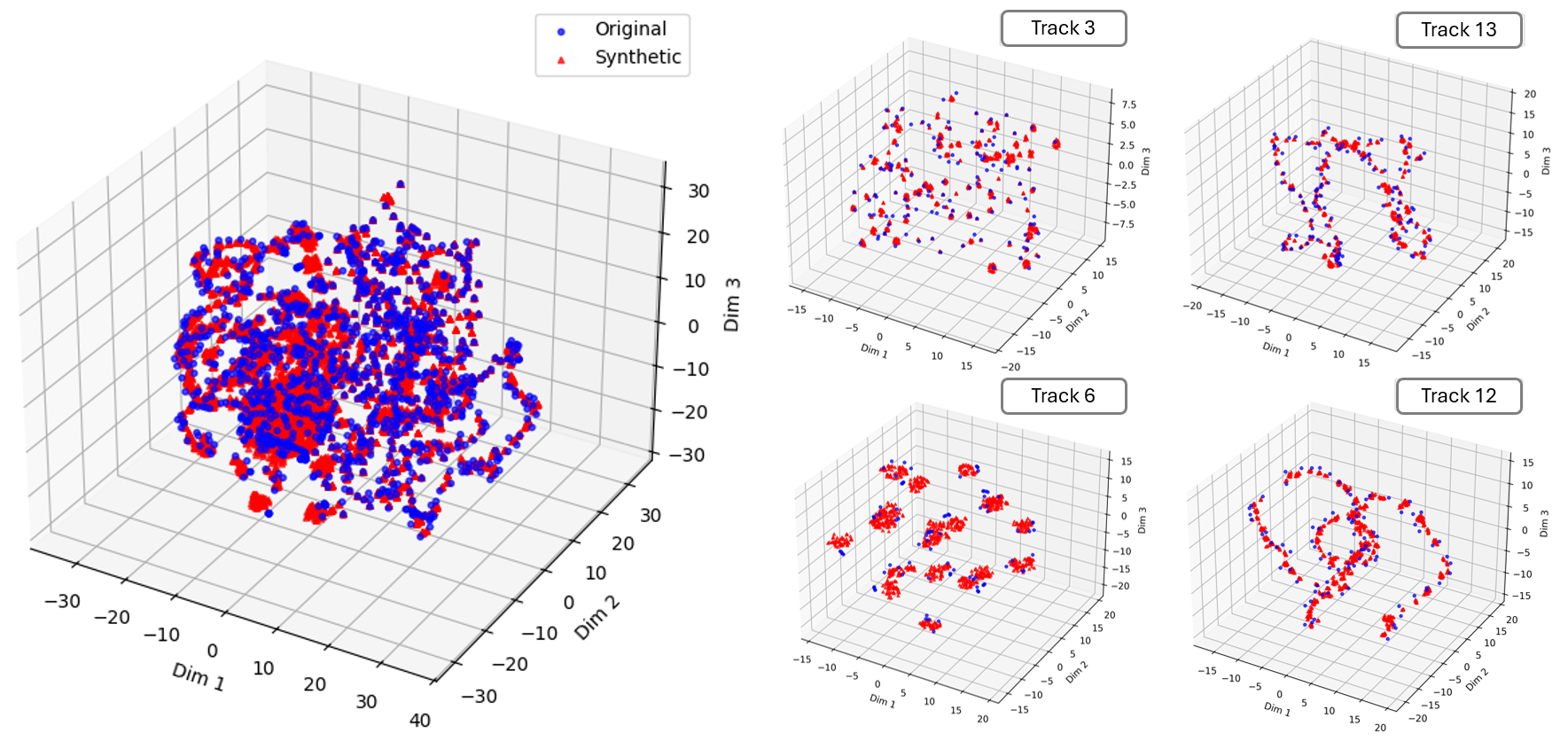}
    \caption{t-SNE projection comparing real and synthetic AoA feature vectors (MUSIC) for the LoS and NLoS regions (dataset size = 1000).}
    \label{fig:CVAE_tsne_light}
    \vspace{-2.5ex}
\end{figure*}

To further examine CVAE behavior, a three dimensional t-SNE visualization is presented in Fig.~\ref{fig:CVAE_tsne_light}, illustrating the overall distribution of real and synthetic data as well as details for track~3, track~13 (NLoS region) and track~6, track~12 (LoS region). CVAE reconstructs 100 synthetic samples per class, resulting in a total of 1000 generated samples. Depending on the underlying distribution of each track, the generated samples cluster consistently with the structure of the original data.

In addition, to assess the quality of the generated samples and their impact on localization performance, we use the tuned random forest classifier which is trained on real data (Section~\ref{sec:framework} Part~\ref{sec:framework:offline_learning}) to predict labels for synthetic samples. Five tracks from LoS region are selected and 1000 samples are generated for each class, resulting in a total of 5000 generated samples. The corresponding classification results are reported in Table~\ref{tbl:CGAN_CVAE_train}. Since the synthetic data generated by the CGAN exhibit a distribution that deviates significantly from the training and test data, the resulting test samples become difficult for the model to predict. Consequently, the performance of the model decreases significantly, reaching an accuracy of approximately 37\%. In contrast, CVAE-based samples introduce only slight deviations from the original data distribution, which better reflect small changes in external conditions. As a result, the model maintains strong performance, achieving an accuracy of approximately 86\%. Based on the experimental results, CVAE demonstrates its suitability as a supporting module within the online learning-based localization framework. In particular, CVAE-generated samples enable effective upsampling of small replay buffers, which is critical for rehearsal and reinitialization processes. In addition, synthetic samples provide a useful tool for evaluating model robustness under distributional variations.

\begin{table}[t]
\caption{Evaluation results of a tuned RF classifier on synthetic AoA samples (LoS region, 5 tracks).}
\label{tbl:CGAN_CVAE_train}
\begin{tabular}{|c|c|c|c|c|}
\hline
\textbf{Model} & \textbf{Acc. (\%)} & \textbf{Precision (\%)} & \textbf{Recall (\%)} & \textbf{F1-score (\%)} \\ \hline
CGAN           & 37.40                  & 24.50                   & 37.40                & 28.38                  \\ \hline
CVAE           & 86.16                  & 88.42                   & 86.16                & 85.67                  \\ \hline
\end{tabular}
\vspace{-2.5ex}
\end{table}


\subsubsection{Batch Retraining with Conventional ML}
\begin{figure*}[t]
    \centering
    \includegraphics[width=0.93\linewidth]{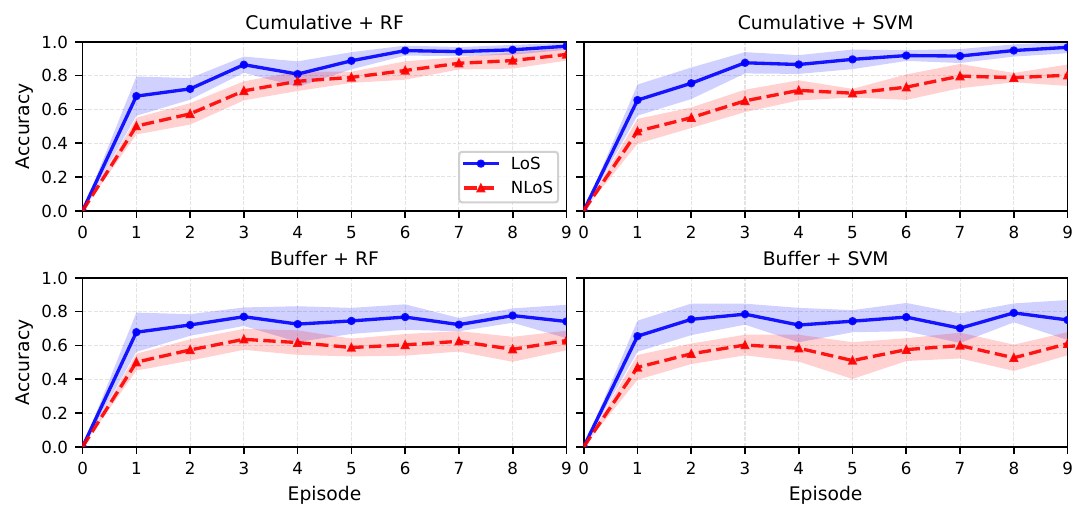}
    \caption{Comparison of buffer and cumulative batch retraining for RF and SVM (mean accuracy over 10 trials is shown, shaded bands denote standard deviation).}
    \label{fig:online_learning_result_1}
    \vspace{-2.5ex}
\end{figure*}

To compare buffer-based and cumulative batch retraining strategies in the offline learning scheme, the dataset is divided into 10 sequential batches, each containing 10\% of the total samples. This setup mimics a deployment scenario in which the offline learning framework periodically retrains the model as new data become available or retrains the model when predefined retraining criteria are met. We evaluate two retraining approaches:
\begin{itemize}
    \item Buffer retraining: the model is retrained using only a recent data window, e.g., the latest 10\% of the dataset.
    \item Cumulative retraining: the model is retrained using all samples observed so far, stored in the feature repository.
\end{itemize}

The experiments are repeated over 10 trials. In Fig.~\ref{fig:online_learning_result_1}, we report the mean accuracy with the corresponding standard deviation of RF and SVM with the two retraining schemes. In general, cumulative retraining achieves higher accuracy as it retains information from previously observed samples. In contrast, buffer retraining creates a wider and more fluctuating standard deviation band, indicating less stable performance over time. Although buffer retraining is more resource-efficient, it may suffer from instability and forgetting when the buffer window is too small.

\subsubsection{Continual Learning for Streaming Data}

In this experiment, six incremental learning models are deployed, including aggregated mondrian forests (AMF), adaptive random forests (ARF), gaussian naïve bayes (GNB), hoeffding adaptive trees (HAT), hoeffding trees (HT), and streaming random patches (SRP). All models are configured using hyperparameter settings adopted from the River benchmark for multiclass classification tasks~\cite{riverml_benchmark}. A confidence threshold $\tau = 0.5$ is applied, which means that only samples with predicted class probability greater than or equal to 50\% are accepted to update the models. To simulate a realistic deployment scenario for the online learning framework, CVAE-based data augmentation is applied to generate 1000 synthetic samples per class. The first 10\% of the dataset is used for warm-up training as the initialization of the model. The remaining 90\% of the data is treated as a streaming sequence, where the AoA feature vectors are processed sequentially and fed into the models one sample at a time.

\textbf{Evaluation metrics.} To ensure a fair comparison, we keep the same AoA feature extraction pipeline for all models and evaluate them under the same training strategies and metrics:
\begin{itemize}
    \item Warm-up accuracy: accuracy measured immediately after the warm-up phase (model initialization).
    \item Online accuracy: prequential accuracy computed over the online phase.
    \item Acceptance rate: we apply a confidence threshold $\tau=0.5$ which means if a confidence probability is over 50\% after a prediction, the sample is accepted and used for learning.
    \item Inference latency and update (training) time: to measure computational cost.
    \item Forgetting rate: to measure the ability of classifier against catastrophic forgetting, we compute the forgetting rate (FR) as defined:
    \begin{equation}
    \text{FR} = \frac{\text{FE}}{T},
    \end{equation}
    where: 
    \begin{equation}
        \text{FE} = \sum_{t=1}^T \mathbbm{1}[ f(x_{t-1})=y_{t-1} \land f(x_t) \neq y_t ],
    \end{equation}
    is the number of forgetting events over time $T$, that means the model made a correct prediction at time $t-1$ but misclassified the sample at time $t$. Lower FR indicates better memory retention. 
\end{itemize}

\begin{figure}
    \centering
    \includegraphics[width=\linewidth]{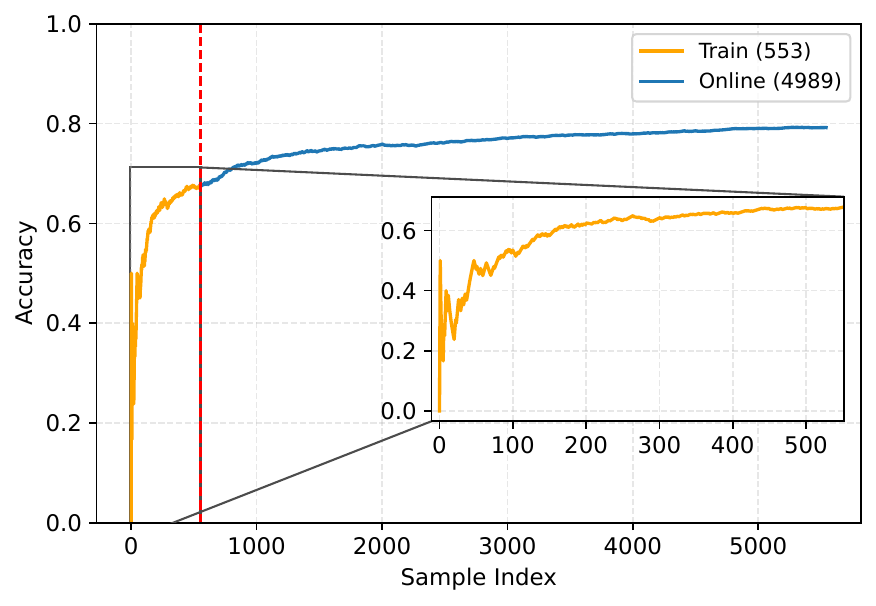}
    \caption{The hoeffding adaptive tree (HAT) classifier peformance of warm-up training and online inference phases applied to AoA features estimated using MUSIC.}
    \label{fig:online_learning_2_details}
    \vspace{-2.5ex}
\end{figure}

Fig.~\ref{fig:online_learning_2_details} illustrates the learning process of the hoeffding adaptive tree classifier applied to AoA features estimated using the MUSIC algorithm. There are two phases: an initial warm-up (model initialization) phase and an online inference and update phase. During the warm-up phase (zoomed region in Fig.~\ref{fig:online_learning_2_details}), the model is exposed to streaming data for the first time and incrementally constructs its decision structure. Due to  the absence of sufficient statistics in the early stage, initial predictions are unreliable, resulting in noticeable accuracy fluctuations and a temporary performance drop within the first 200 samples. As more observations become available, the classification accuracy progressively increases and stabilizes once it reaches approximately 70\%. In the online inference phase, the model continues to refine its decision boundaries in an incremental approach, leading to a gradual performance improvement that reaches 77.75\% accuracy, without any sudden degradation or catastrophic collapse. This behavior is inherent to hoeffding-based decision trees, which require a sufficient number of samples to statistically validate a split. Until the hoeffding bound is satisfied, candidate splits remain uncertain and sample routing may be suboptimal. Once the bound is met, confident splits are performed, achieving accuracy recovery. After sufficient splits and leaf updates, the model converges to a more stable decision structure, resulting in consistent and robust predictions over time.

The incremental learning results for six incremental models under both LoS and NLoS conditions are reported in Table~\ref{tbl:online_result_2}. Across all scenarios (LoS / NLoS, MUSIC / ESPRIT), AMF consistently achieves the best online performance, reaching online accuracy around 0.94 while maintaining the lowest forgetting rate (0.025-0.043). This indicates strong stability under the prequential predict-then-train strategy and shows that AMF adapts effectively without frequent degradation over time. SRP is the second-best online accuracy in both regions. ESPRIT yields a clear improvement over MUSIC in NLoS (0.753 vs 0.728), implying that the AoA estimator can meaningfully affect online learning robustness in more challenging propagation conditions. In contrast, GaussianNB and Hoeffding-tree variants accept almost all samples (acceptance is approximately 100\%) achieve lower online accuracy and higher forgetting rates, suggesting that frequent incremental updates may lead to unstable decision boundaries rather than consistent improvement.

\begin{table}[t]
\caption{Computational cost comparison between continual tree-based and ensemble classifiers and the continual ProtoNet-based few-shot approach. Results are averaged across LoS / NLoS and MUSIC / ESPRIT configurations.}
\label{tbl:online_continual_protonet_timing}
\centering
\begin{tabular}{|c|cc|cc|}
\hline
\textbf{Model}  & \multicolumn{2}{c|}{\textbf{Mean (ms)}}                   & \multicolumn{2}{c|}{\textbf{Total (s)}}                   \\ \hline
\textbf{}       & \multicolumn{1}{c|}{\textbf{Inference}} & \textbf{Update} & \multicolumn{1}{c|}{\textbf{Inference}} & \textbf{Update} \\ \hline
AMF             & \multicolumn{1}{c|}{1.6925}             & 17.7248         & \multicolumn{1}{c|}{8.4868}             & 88.86           \\ \hline
ARF             & \multicolumn{1}{c|}{1.6048}             & 4.28            & \multicolumn{1}{c|}{8.048}              & 21.4613         \\ \hline
GNB             & \multicolumn{1}{c|}{2.1538}             & 0.4258          & \multicolumn{1}{c|}{10.7988}            & 2.1355          \\ \hline
HAT             & \multicolumn{1}{c|}{2.3275}             & 2.7475          & \multicolumn{1}{c|}{11.67}              & 13.7763         \\ \hline
HT              & \multicolumn{1}{c|}{2.2795}             & 1.6395          & \multicolumn{1}{c|}{11.429}             & 8.2205          \\ \hline
SRP             & \multicolumn{1}{c|}{10.9113}            & 77.398          & \multicolumn{1}{c|}{54.7395}            & 388.1073        \\ \hline
ProtoNet (k=1)  & \multicolumn{1}{c|}{1.05}               & 4.775           & \multicolumn{1}{c|}{0.9703}             & 4.3793          \\ \hline
ProtoNet (k=5)  & \multicolumn{1}{c|}{1.125}              & 5.05            & \multicolumn{1}{c|}{0.2545}             & 1.1415          \\ \hline
ProtoNet (k=10) & \multicolumn{1}{c|}{1.225}              & 5.6             & \multicolumn{1}{c|}{0.1563}             & 0.7175          \\ \hline
\end{tabular}
\vspace{-2.5ex}
\end{table}

Table~\ref{tbl:online_continual_protonet_timing} demonstrates that incremental tree-based classifiers maintain millisecond-level inference and update latency, supporting real-time deployment. Among them, AMF and ARF provide a balanced trade-off between accuracy and computational cost, while GNB and HT offer the lowest update overhead at the expense of reduced accuracy. In contrast, ensemble methods such as SRP significantly increase computational cost, with a total update time exceeding 380 seconds, limiting scalability for long streaming sequences.

\begin{figure*}
    \centering
    \includegraphics[width=0.9\linewidth]{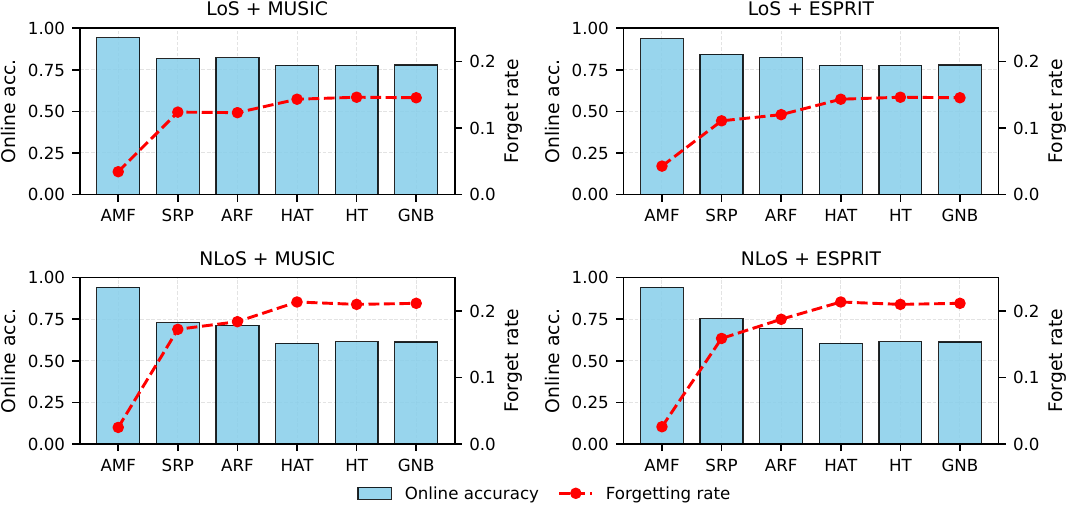}
    \caption{Online Accuracy (column chart) and Forgetting Rate (line chart) comparison across six incremental models in four scenarios (LoS / NLoS $\times$ MUSIC / ESPRIT).}
    \label{fig:online_learning_result_2}
    \vspace{-2ex}
\end{figure*}

Fig.~\ref{fig:online_learning_result_2} summarizes the performance of six incremental learning models. The bar chart reports the online accuracy, where higher values indicate better predictive performance, while the line chart reports the forgetting rate, where lower values indicate stronger retention and reduced catastrophic forgetting. The performance of the six incremental models, including accuracy after warm-up training, accuracy after online learning phase, acceptance rate, and forgetting rate are presented in the Appendix (Table~\ref{tbl:online_result_2_details} and Table~\ref{tbl:online_result_2_details1}).

\begin{table*}[h]
\centering
\caption{Online performance: higher Accuracy (Acc) is better, while a lower Forgetting Rate (FR) is better.
}
\label{tbl:online_result_2}
\begin{tabular}{|cc|cc|cc|cc|cc|cc|cc|}
\hline
\multicolumn{2}{|c|}{\multirow{2}{*}{\textbf{}}}                       & \multicolumn{2}{c|}{\textbf{AMF}}               & \multicolumn{2}{c|}{\textbf{SRP}}               & \multicolumn{2}{c|}{\textbf{ARF}}               & \multicolumn{2}{c|}{\textbf{GNB}}               & \multicolumn{2}{c|}{\textbf{HAT}}               & \multicolumn{2}{c|}{\textbf{HT}}                \\ \cline{3-14} 
\multicolumn{2}{|c|}{}                                                 & \multicolumn{1}{c|}{\textbf{Acc}} & \textbf{FR} & \multicolumn{1}{c|}{\textbf{Acc}} & \textbf{FR} & \multicolumn{1}{c|}{\textbf{Acc}} & \textbf{FR} & \multicolumn{1}{c|}{\textbf{Acc}} & \textbf{FR} & \multicolumn{1}{c|}{\textbf{Acc}} & \textbf{FR} & \multicolumn{1}{c|}{\textbf{Acc}} & \textbf{FR} \\ \hline
\multicolumn{1}{|c|}{\multirow{2}{*}{\textbf{LoS}}}  & \textbf{MUSIC}  & \multicolumn{1}{c|}{0.9432}       & 0.0343      & \multicolumn{1}{c|}{0.8161}       & 0.1239      & \multicolumn{1}{c|}{0.823}        & 0.1231      & \multicolumn{1}{c|}{0.779}        & 0.1455      & \multicolumn{1}{c|}{0.7775}       & 0.1433      & \multicolumn{1}{c|}{0.7782}       & 0.1463      \\ \cline{2-14} 
\multicolumn{1}{|c|}{}                               & \textbf{ESPRIT} & \multicolumn{1}{c|}{0.9383}       & 0.0427      & \multicolumn{1}{c|}{0.8423}       & 0.1108      & \multicolumn{1}{c|}{0.8244}       & 0.1201      & \multicolumn{1}{c|}{0.779}        & 0.1455      & \multicolumn{1}{c|}{0.7775}       & 0.1433      & \multicolumn{1}{c|}{0.7782}       & 0.1463      \\ \hline
\multicolumn{1}{|c|}{\multirow{2}{*}{\textbf{NLoS}}} & \textbf{MUSIC}  & \multicolumn{1}{c|}{0.9409}       & 0.0248      & \multicolumn{1}{c|}{0.7282}       & 0.1723      & \multicolumn{1}{c|}{0.713}        & 0.1838      & \multicolumn{1}{c|}{0.6135}       & 0.2114      & \multicolumn{1}{c|}{0.604}        & 0.2134      & \multicolumn{1}{c|}{0.6153}       & 0.2098      \\ \cline{2-14} 
\multicolumn{1}{|c|}{}                               & \textbf{ESPRIT} & \multicolumn{1}{c|}{0.9412}       & 0.0258      & \multicolumn{1}{c|}{0.7534}       & 0.1586      & \multicolumn{1}{c|}{0.6962}       & 0.1874      & \multicolumn{1}{c|}{0.6135}       & 0.2114      & \multicolumn{1}{c|}{0.604}        & 0.2134      & \multicolumn{1}{c|}{0.6153}       & 0.2098      \\ \hline
\end{tabular}
\end{table*}

\subsection{FEW-SHOT LEARNING}
\label{sec:experiments:few_shot}

\subsubsection{Prototypical Networks - Standard Episodic Meta-learning}
\textbf{Experiment setup.} We evaluate standard ProtoNet for few-shot track identification separately under LoS and NLoS conditions. The AoA feature vectors are derived from CSI using MUSIC and ESPRIT algorithms. Each feature vector is computed over a window of 2000 snapshots, as described in Section~\ref{sec:experiments} Part~\ref{sec:experiments:feature_extraction}. The LoS set includes tracks $\{6, 9, 10, 11, 12\}$ and NLoS set includes tracks $\{1, 2, 3, 13, 20\}$. For each region, we construct $N-way$ $K-shot$ episodes with $K \in \{1,2,4,8,12\}$ and $N=3$. Using $N=3$ allows us to simulate the case of insufficient samples, only a small number of classes and labeled examples are available. ProtoNet consists of an embedding network $f_\theta$ implemented as a 3-layer MLP as described in Table~\ref{tbl:protonet}. For each $(N, K)$ episode, experiments are repeated over 10 trials to calculate the mean accuracy with 95\% confidence interval.

We train ProtoNet for 1000 meta-training episodes using Adam with a learning rate $10^{-3}$. Evaluation is performed on 200 meta-test episodes sampled from the held-out split. We use a standard 80 / 20 split of the available samples within each region to form meta-train and meta-test sets prior to episodic sampling, so meta-test can be simulated the case of unseen samples (potentially different track subsets depending on episode construction) within the same region.
Table~\ref{tbl:protonet_result_1} summarizes the few-shot results in terms of mean episodic accuracy and 95\% confidence intervals. Across both LoS and NLoS regions, performance improves as the number of shots $K$ increases, indicating that ProtoNet benefits from additional labeled support examples per track. Under LoS, ProtoNet achieves strong performance even with low shot, improving accuracy from 0.7898 $\pm$ 0.0069 (1-shot) to 0.9029 $\pm$ 0.0111 (4-shot) and then saturating around 0.914-0.915 for $K\ge8$. In contrast, NLoS remains more challenging: accuracy increases from 0.5466 $\pm$ 0.0166 (1-shot) to 0.7566 $\pm$ 0.0115 (12-shot), with wider confidence intervals, reflecting greater variability across trials. Overall, these results confirm that additional support examples improve performance in both cases and the persistent LoS-NLoS gap suggests that NLoS AoA features shows the difficulty in processing from the NLoS region due to multipath effects.

\begin{table}[h]
\caption{LoS vs NLoS comparison (ProtoNet, MUSIC, N=3, mean accuracy with 95\% CI) and LoS-NLoS w.r.t $K$.}
\label{tbl:protonet_result_1}
\centering
\begin{tabular}{|c|c|c|c|}
\hline
\textbf{K-shot} & \textbf{\begin{tabular}[c]{@{}c@{}}LoS \\ acc $\pm$ 95\% CI\end{tabular}} & \textbf{\begin{tabular}[c]{@{}c@{}}NLoS \\ acc $\pm$ 95\% CI\end{tabular}} & \textbf{\begin{tabular}[c]{@{}c@{}}Gap \\ (LoS - NLoS)\end{tabular}} \\ \hline
1               & 0.7898 $\pm$ 0.0069                                                            & 0.5466 $\pm$ 0.0166                                                             & 0.2432                                                               \\ \hline
2               & 0.8541 $\pm$ 0.0145                                                            & 0.6803 $\pm$ 0.0215                                                             & 0.1738                                                               \\ \hline
4               & 0.9029 $\pm$ 0.0111                                                            & 0.7250 $\pm$ 0.0146                                                             & 0.1779                                                               \\ \hline
8               & 0.9135 $\pm$ 0.0123                                                            & 0.7500 $\pm$ 0.0159                                                             & 0.1635                                                               \\ \hline
12              & 0.9148 $\pm$ 0.0136                                                            & 0.7566 $\pm$ 0.0115                                                             & 0.1582                                                               \\ \hline
\end{tabular}
\vspace{-2ex}
\end{table}

\subsubsection{Continual Few-shot Learning via ProtoNet}

In this experiment, we leverage CVAE-based data augmentation to expand the dataset to 1000 samples per class. We then construct a sequence of $N-way$ $K-shot$ tasks from these samples, each episode represents a subset of the newly available data and samples are not reused across episodes, in contrast to standard episodic meta-training. Each episode uses $N=3$ classes, with $K$ support samples per class and $Q$ query samples per class.

\textbf{Evaluation metrics.} The performance of continual few-shot learning is assessed using classification accuracy, episode-level forgetting rate to quantify catastrophic forgetting, and computational cost measured through inference latency and update (training) time. Let $E$ denote the number of episodes obtained from the split, an episode $e$ consists of a support set and a query set, and episodes arrive sequentially as $e = 1, \dots, E$. The model is updated after each episode. The forgetting rate $FR(e)$ of episode $e$ is defined as the performance degradation in previously learned episodes caused by updating the model with episode $e$.

In continual learning with streaming data, forgetting rate captures short-term instability as data arrive sequentially over time. In contrast, for prototypical networks operating in a continual few-shot setting, forgetting is defined at the episode level and measures the degradation of performance on previously observed episodes after sequential updates. For episode $e \geqslant 2$, the episode-level forgetting rate is defined as:
\begin{equation}
\text{FR}(e) = \frac{1}{e-1}
\sum_{i=1}^{e-1}
\max\Big( 0, \text{acc}_i^{\text{before}}(e)
-
\text{acc}_i^{\text{after}}(e) \Big)    
\end{equation}
where:
\begin{itemize}
    \item $\text{acc}_i^{\text{before}}(e)$ denotes the classification accuracy in episode $i$ before updating the model with episode $e$,
    \item $\text{acc}_i^{\text{after}}(e)$ denotes the classification accuracy in episode $i$ after updating the model with episode $e$.
\end{itemize}

\begin{figure}[t]
    \centering
    \includegraphics[width=0.95\linewidth]{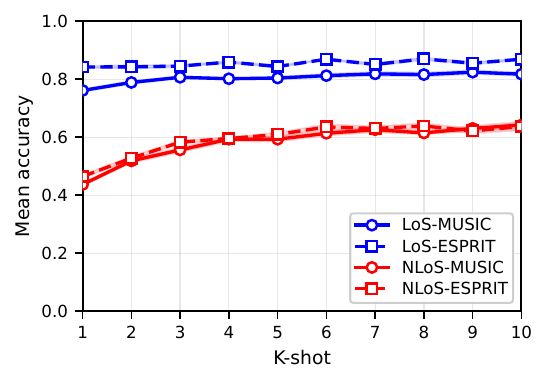}
    \caption{Classification accuracy (mean $\pm$ 95\% confidence interval) w.r.t $K$ for continual few-shot learning under four scenarios (LoS / NLoS $\times$ MUSIC / ESPRIT).}
    \label{fig:protonet_online_result}
    \vspace{-2ex}
\end{figure}

Table~\ref{tbl:protonet_result_2} presents the accuracies for different values of $K$ and its corresponding valid episodes $E$ over 10 trials. Overall, LoS consistently outperforms NLoS, while in the NLoS setting, both AoA estimators improve their performance with $K$, LoS results are more stable across $K$. Comparing AoA estimators, ESPRIT achieves higher final accuracy than MUSIC under LoS condition (increasing from 0.8425 ($K=1$) to 0.8693 ($K=10$). The number of valid episodes decreases for larger $K$ because each episode requires more labeled support samples per class, thus, the reported accuracies correspond to the last episode in each configuration. Fig.~\ref{fig:protonet_online_result} illustrates the final classification accuracy with respect to $K$ ($K = 1, \ldots, 10$) in both LoS and NLoS regions, using AoA features extracted via the MUSIC and ESPRIT algorithms.

\begin{table*}[t]
\footnotesize
\caption{Final accuracy (mean $\pm$ 95\% CI) vs $K$. Number of valid episodes $E$ for ProtoNet episodic sampling under $N$-way $K$-shot $Q$-query settings (LoS / NLoS; MUSIC / ESPRIT).}
\label{tbl:protonet_result_2}
\centering
\begin{tabular}{|cc|cccccc|cccccc|}
\hline
\multicolumn{2}{|c|}{\multirow{2}{*}{\textbf{}}}  & \multicolumn{6}{c|}{\textbf{LoS}}                                                                                                                                                                                                                                                                                  & \multicolumn{6}{c|}{\textbf{NLoS}}                                                                                                                                                                                                                                                                                 \\ \cline{3-14} 
\multicolumn{2}{|c|}{}                            & \multicolumn{3}{c|}{\textbf{MUSIC}}                                                                                                                                & \multicolumn{3}{c|}{\textbf{ESPRIT}}                                                                                                          & \multicolumn{3}{c|}{\textbf{MUSIC}}                                                                                                                                & \multicolumn{3}{c|}{\textbf{ESPRIT}}                                                                                                          \\ \hline
\multicolumn{1}{|c|}{\textbf{$K$}} & \textbf{$N = Q$} & \multicolumn{1}{c|}{\textbf{$E$}} & \multicolumn{1}{c|}{\textbf{\begin{tabular}[c]{@{}c@{}}mean \\ $\pm$ 95\% CI\end{tabular}}} & \multicolumn{1}{c|}{\textbf{FR}} & \multicolumn{1}{c|}{\textbf{$E$}} & \multicolumn{1}{c|}{\textbf{\begin{tabular}[c]{@{}c@{}}mean \\ $\pm$ 95\% CI\end{tabular}}} & \textbf{FR} & \multicolumn{1}{c|}{\textbf{$E$}} & \multicolumn{1}{c|}{\textbf{\begin{tabular}[c]{@{}c@{}}mean \\ $\pm$ 95\% CI\end{tabular}}} & \multicolumn{1}{c|}{\textbf{FR}} & \multicolumn{1}{c|}{\textbf{$E$}} & \multicolumn{1}{c|}{\textbf{\begin{tabular}[c]{@{}c@{}}mean \\ $\pm$ 95\% CI\end{tabular}}} & \textbf{FR} \\ \hline
\multicolumn{1}{|c|}{1}            & 3            & \multicolumn{1}{c|}{906}          & \multicolumn{1}{c|}{\begin{tabular}[c]{@{}c@{}}0.7616 \\ $\pm$ 0.0037\end{tabular}}         & \multicolumn{1}{c|}{0.0124}      & \multicolumn{1}{c|}{909}          & \multicolumn{1}{c|}{\begin{tabular}[c]{@{}c@{}}0.8425 \\ $\pm$ 0.0031\end{tabular}}         & 0.0080       & \multicolumn{1}{c|}{918}          & \multicolumn{1}{c|}{\begin{tabular}[c]{@{}c@{}}0.4371\\ $\pm$ 0.0040\end{tabular}}          & \multicolumn{1}{c|}{0.0222}      & \multicolumn{1}{c|}{922}          & \multicolumn{1}{c|}{\begin{tabular}[c]{@{}c@{}}0.4649\\ $\pm$ 0.0052\end{tabular}}          & 0.0218      \\ \hline
\multicolumn{1}{|c|}{2}            & 3            & \multicolumn{1}{c|}{452}          & \multicolumn{1}{c|}{\begin{tabular}[c]{@{}c@{}}0.7892 \\ $\pm$ 0.0051\end{tabular}}         & \multicolumn{1}{c|}{0.0076}      & \multicolumn{1}{c|}{451}          & \multicolumn{1}{c|}{\begin{tabular}[c]{@{}c@{}}0.8431 \\ $\pm$ 0.0057\end{tabular}}         & 0.0051      & \multicolumn{1}{c|}{456}          & \multicolumn{1}{c|}{\begin{tabular}[c]{@{}c@{}}0.5186\\ $\pm$ 0.0063\end{tabular}}          & \multicolumn{1}{c|}{0.0138}      & \multicolumn{1}{c|}{454}          & \multicolumn{1}{c|}{\begin{tabular}[c]{@{}c@{}}0.5280\\ $\pm$ 0.0064\end{tabular}}          & 0.0137      \\ \hline
\multicolumn{1}{|c|}{3}            & 3            & \multicolumn{1}{c|}{302}          & \multicolumn{1}{c|}{\begin{tabular}[c]{@{}c@{}}0.8070 \\ $\pm$ 0.0037\end{tabular}}         & \multicolumn{1}{c|}{0.0067}      & \multicolumn{1}{c|}{298}          & \multicolumn{1}{c|}{\begin{tabular}[c]{@{}c@{}}0.8455 \\ $\pm$ 0.0036\end{tabular}}         & 0.0046      & \multicolumn{1}{c|}{304}          & \multicolumn{1}{c|}{\begin{tabular}[c]{@{}c@{}}0.5565\\ $\pm$ 0.0086\end{tabular}}          & \multicolumn{1}{c|}{0.0104}      & \multicolumn{1}{c|}{306}          & \multicolumn{1}{c|}{\begin{tabular}[c]{@{}c@{}}0.5832\\ $\pm$ 0.0055\end{tabular}}          & 0.0109      \\ \hline
\multicolumn{1}{|c|}{4}            & 3            & \multicolumn{1}{c|}{256}          & \multicolumn{1}{c|}{\begin{tabular}[c]{@{}c@{}}0.8020 \\ $\pm$ 0.0026\end{tabular}}         & \multicolumn{1}{c|}{0.0061}      & \multicolumn{1}{c|}{256}          & \multicolumn{1}{c|}{\begin{tabular}[c]{@{}c@{}}0.8595 \\ $\pm$ 0.0033\end{tabular}}         & 0.0043      & \multicolumn{1}{c|}{262}          & \multicolumn{1}{c|}{\begin{tabular}[c]{@{}c@{}}0.5934\\ $\pm$ 0.0072\end{tabular}}          & \multicolumn{1}{c|}{0.0100}        & \multicolumn{1}{c|}{261}          & \multicolumn{1}{c|}{\begin{tabular}[c]{@{}c@{}}0.5959\\ $\pm$ 0.0040\end{tabular}}          & 0.0104      \\ \hline
\multicolumn{1}{|c|}{5}            & 3            & \multicolumn{1}{c|}{224}          & \multicolumn{1}{c|}{\begin{tabular}[c]{@{}c@{}}0.8045 \\ $\pm$ 0.0040\end{tabular}}         & \multicolumn{1}{c|}{0.0071}      & \multicolumn{1}{c|}{226}          & \multicolumn{1}{c|}{\begin{tabular}[c]{@{}c@{}}0.8442 \\ $\pm$ 0.0036\end{tabular}}         & 0.0044      & \multicolumn{1}{c|}{224}          & \multicolumn{1}{c|}{\begin{tabular}[c]{@{}c@{}}0.5934\\ $\pm$ 0.0086\end{tabular}}          & \multicolumn{1}{c|}{0.0092}      & \multicolumn{1}{c|}{227}          & \multicolumn{1}{c|}{\begin{tabular}[c]{@{}c@{}}0.6098\\ $\pm$ 0.0063\end{tabular}}          & 0.0094      \\ \hline
\multicolumn{1}{|c|}{6}            & 3            & \multicolumn{1}{c|}{179}          & \multicolumn{1}{c|}{\begin{tabular}[c]{@{}c@{}}0.8127 \\ $\pm$ 0.0063\end{tabular}}         & \multicolumn{1}{c|}{0.0057}      & \multicolumn{1}{c|}{179}          & \multicolumn{1}{c|}{\begin{tabular}[c]{@{}c@{}}0.8693 \\ $\pm$ 0.0014\end{tabular}}         & 0.0040       & \multicolumn{1}{c|}{183}          & \multicolumn{1}{c|}{\begin{tabular}[c]{@{}c@{}}0.6138\\ $\pm$ 0.0076\end{tabular}}          & \multicolumn{1}{c|}{0.0082}      & \multicolumn{1}{c|}{180}          & \multicolumn{1}{c|}{\begin{tabular}[c]{@{}c@{}}0.6357\\ $\pm$ 0.0090\end{tabular}}          & 0.0085      \\ \hline
\multicolumn{1}{|c|}{7}            & 3            & \multicolumn{1}{c|}{162}          & \multicolumn{1}{c|}{\begin{tabular}[c]{@{}c@{}}0.8188 \\ $\pm$ 0.0066\end{tabular}}         & \multicolumn{1}{c|}{0.0061}      & \multicolumn{1}{c|}{164}          & \multicolumn{1}{c|}{\begin{tabular}[c]{@{}c@{}}0.8513 \\ $\pm$ 0.0045\end{tabular}}         & 0.0042      & \multicolumn{1}{c|}{165}          & \multicolumn{1}{c|}{\begin{tabular}[c]{@{}c@{}}0.6262\\ $\pm$ 0.0090\end{tabular}}          & \multicolumn{1}{c|}{0.0077}      & \multicolumn{1}{c|}{164}          & \multicolumn{1}{c|}{\begin{tabular}[c]{@{}c@{}}0.6305\\ $\pm$ 0.0083\end{tabular}}          & 0.0083      \\ \hline
\multicolumn{1}{|c|}{8}            & 3            & \multicolumn{1}{c|}{151}          & \multicolumn{1}{c|}{\begin{tabular}[c]{@{}c@{}}0.8164 \\ $\pm$ 0.0072\end{tabular}}         & \multicolumn{1}{c|}{0.0059}      & \multicolumn{1}{c|}{148}          & \multicolumn{1}{c|}{\begin{tabular}[c]{@{}c@{}}0.8705 \\ $\pm$ 0.0043\end{tabular}}         & 0.0046      & \multicolumn{1}{c|}{150}          & \multicolumn{1}{c|}{\begin{tabular}[c]{@{}c@{}}0.6150\\ $\pm$ 0.0054\end{tabular}}          & \multicolumn{1}{c|}{0.0082}      & \multicolumn{1}{c|}{149}          & \multicolumn{1}{c|}{\begin{tabular}[c]{@{}c@{}}0.6390\\ $\pm$ 0.0092\end{tabular}}          & 0.0082      \\ \hline
\multicolumn{1}{|c|}{9}            & 3            & \multicolumn{1}{c|}{137}          & \multicolumn{1}{c|}{\begin{tabular}[c]{@{}c@{}}0.8247 \\ $\pm$ 0.0061\end{tabular}}         & \multicolumn{1}{c|}{0.0064}      & \multicolumn{1}{c|}{137}          & \multicolumn{1}{c|}{\begin{tabular}[c]{@{}c@{}}0.8555 \\ $\pm$ 0.0044\end{tabular}}         & 0.0050       & \multicolumn{1}{c|}{140}          & \multicolumn{1}{c|}{\begin{tabular}[c]{@{}c@{}}0.6307\\ $\pm$ 0.0093\end{tabular}}          & \multicolumn{1}{c|}{0.0073}      & \multicolumn{1}{c|}{136}          & \multicolumn{1}{c|}{\begin{tabular}[c]{@{}c@{}}0.6219\\ $\pm$ 0.0083\end{tabular}}          & 0.0087      \\ \hline
\multicolumn{1}{|c|}{10}           & 3            & \multicolumn{1}{c|}{126}          & \multicolumn{1}{c|}{\begin{tabular}[c]{@{}c@{}}0.8177\\ $\pm$ 0.0058\end{tabular}}          & \multicolumn{1}{c|}{0.0055}      & \multicolumn{1}{c|}{125}          & \multicolumn{1}{c|}{\begin{tabular}[c]{@{}c@{}}0.8693 \\ $\pm$ 0.0028\end{tabular}}         & 0.0040       & \multicolumn{1}{c|}{128}          & \multicolumn{1}{c|}{\begin{tabular}[c]{@{}c@{}}0.6424\\ $\pm$ 0.0109\end{tabular}}          & \multicolumn{1}{c|}{0.0076}      & \multicolumn{1}{c|}{129}          & \multicolumn{1}{c|}{\begin{tabular}[c]{@{}c@{}}0.6373\\ $\pm$ 0.0135\end{tabular}}          & 0.0081      \\ \hline
\end{tabular}
\end{table*}

\begin{figure*}[!h]
    \centering
    \includegraphics[width=0.95\linewidth]{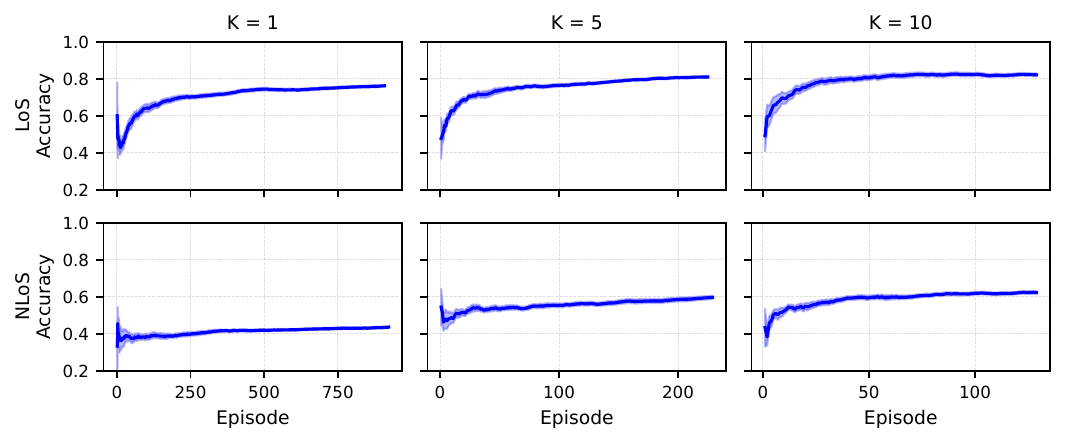}
    \caption{Evolution of the mean accuracy over training episodes for ProtoNet under three $K$-shot settings ($K=1, 5, 10$; MUSIC). Solid lines indicate the mean accuracy, while the shaded regions correspond to the 95\% confidence intervals.}
    \label{fig:protonet_online_evolution}
    \vspace{-2ex}
\end{figure*}
Fig.~\ref{fig:protonet_online_evolution} reports the evolution of the mean accuracy over training episodes, averaged over 10 trials (95\% CI) for three representative values $K$ ($K=1, 5, 10$), AoA features extracted using the MUSIC algorithm. Across all conditions, accuracy increases rapidly in the early episodes and then improves more gradually, indicating progressive stabilization of the embedding space and the prototype estimates.

Across four cases presented in Table~\ref{tbl:protonet_result_2}, we observe that increasing support size $K$ improves representation stability and reduces catastrophic forgetting. With a larger number of support samples, the network can compute more reliable prototype estimates, resulting in less noisy embedding updates. For example, in the NLoS + MUSIC configuration, the forgetting rate decreases from 0.0222 at $K=1$ to 0.0076 at $K=10$, corresponding to approximately a  reduction of 65.8\%. Moreover, NLoS region consistently has higher forgetting rates than LoS region. This is due to variable and unstable AoA representations, making episodic updates more disruptive to previously learned knowledge. For example, at $K=1$, the forgetting rate for MUSIC and ESPRIT under NLoS (0.0222 and 0.0218, respectively) is higher than those under LoS (0.0124 and 0.0080, respectively). In addition, forgetting rates decrease significantly from $K=1$ to $K=4$ and stabilize for $K\ge 5$. A similar stabilization trend is observed in classification accuracy, with no substantial performance gains beyond moderate support sizes. 

Regarding the inference and update times of this method, Table~\ref{tbl:online_continual_protonet_timing} presents three representative results for $K=1, 5, 10$. The mean inference time of ProtoNet remains within 1.05-1.23 ms per episode. Similarly, the mean update time remains approximately 5 ms per episode. Although larger support sizes $K$ slightly increase per-episode computational cost, they reduce the total number of episodes required, resulting in a shorter total update time. These findings indicate that continual ProtoNet-based few-shot learning with a moderate support size is sufficient to stabilize continual few-shot learning while avoiding unnecessary computational overhead.

\subsection{Discussion}
This section presented the experimental setup and results for AoA-based localization frameworks under both offline and online learning settings. In the offline learning framework, the fine-tuned models achieve high localization performance, with accuracy reaching up to 99.8\% for trajectory classification in both LoS and NLoS regions. These results indicate that offline learning can provide a strong and stable baseline, supporting its use as a component in physical-layer authentication or as an input feature for trust evaluation schemes, as proposed in~\cite{trinh-mnl2024}.

For the online learning framework, we investigate two learning approaches in simulated dynamic environments generated using CVAE-based data augmentation. The first approach employs incremental learning with tree-based classifiers and ensemble models to continuously update the model  under streaming data conditions. The second approach explores continual few-shot learning using prototypical networks, enabling rapid adaptation when only limited labeled samples are available.

Given the same number of learning samples, the incremental learning approach achieves higher accuracy than the few-shot approach, with average accuracies of 94.32\% and 87.05\%, respectively. In addition, the computational comparison reported in Table~\ref{tbl:online_continual_protonet_timing} reveals distinct online learning strategies between these two approaches. ProtoNet achieves an inference latency in the range of 1.05-1.23 ms per episode, comparable to incremental tree-based models and significantly faster than ensemble-heavy methods such as SRP. ProtoNet also achieves moderate update cost (approximately 5 ms per episode), performs substantially fewer updates than incremental classifiers, resulting in a lower total adaptation time for moderate support sizes (0.72 s for $K=10$). Incremental models update after every incoming sample (about 5000 updates in our setting), enabling immediate adaptation without requiring data buffering. In contrast, ProtoNet aggregates the samples into episodes (900 episodes for $K=1$, about 120 episodes for $K=10$) and performs fewer but heavier updates. While this reduces the total update (training) time for larger support sizes, it introduces a delay due to episode construction and temporary data storage.

In summary, our findings demonstrate that both incremental classifiers and continual ProtoNet-based few-shot learning achieve millisecond-level inference and update costs, making them suitable for real-time or near real-time high-accuracy localization frameworks. Moreover, the consistently strong localization performance achieved under LoS conditions suggests that channel propagation characteristics can be naturally incorporated into trustworthiness assessment. In particular, localization decisions obtained in LoS environments can be treated as more reliable and can serve as trustworthiness metrics for wireless communication tasks.
\section{CONCLUSIONS}
\label{sec:conclusion}

This paper proposed an adaptive AoA-based localization frameworks for wireless environment, consisting of offline learning and online learning strategies to support different deployment conditions. Experimental results on a real-world outdoor mMIMO OFDM CSI dataset showed that the proposed framework can achieve strong localization performance and supports incremental model updates for streaming data. In general, the findings highlight the potential of AoA-based learning for practical low-latency localization. There are several directions for future work. First, further optimization and fine-tuning of continual learning models can be investigated to improve performance and stabilize the learning process. Second, the framework could be extended to more challenging settings to evaluate its robustness under practical impairments such as blockage, calibration mismatch, and interference. Third, the framework could be integrated into physical layer authentication systems or incorporated into trust and reputation management frameworks as a physical layer trust indicator.

\appendices
\section*{APPENDIX}
\label{sec:appendix}



\begin{table}[!h]
\caption{Hyperparameter tuning setup and search spaces}
\label{tbl:searchspace}
\centering
\begin{tabular}{|l|p{6cm}|}
\hline
\textbf{Model} & \textbf{Tuned hyperparameters and their search space} \\
\hline
\multirow{3}{*}{LR} & 
$C \sim \mathrm{LogUniform}[10^{-5},10^{2}]$; \\
& solver $\in \{\mathrm{liblinear}, \mathrm{lbfgs}, \mathrm{saga}\}$; \\
&penalty=l2; max\_iter=1000 \\
\hline
\multirow{3}{*}{KNN} & 
$k \in [1,30]$; weights $\in \{\mathrm{uniform}, \mathrm{distance}\}$;\\ & algorithm $\in \{\mathrm{auto}, \mathrm{ball\_tree}, \mathrm{kd\_tree}, \mathrm{brute}\}$; \\
&leaf\_size $\in [10,50]$ \\
\hline
\multirow{4}{*}{SVM} & 
kernel $\in \{\mathrm{linear}, \mathrm{rbf}, \mathrm{poly}, \mathrm{sigmoid}\}$; \\
&$C \sim \mathrm{LogUniform}[10^{-5},10^{2}]$; \\
& $\gamma \sim \mathrm{LogUniform}[10^{-2},1]$ if non-linear; \\
&degree $\in [2,5]$ if poly \\
\hline
\multirow{4}{*}{RF} & 
n\_estimators $\in [50,300]$; max\_depth $\in [3,30]$; \\
&max\_features $\in \{\mathrm{sqrt}, \mathrm{log2}, \mathrm{None}\}$; \\ 
&min\_samples\_split $\in [2,10]$; \\
&min\_samples\_leaf $\in [1,10]$ \\
\hline
\multirow{4}{*}{GBM} &
n\_estimators $\in [50,300]$; \\
&learning\_rate $\sim \mathrm{LogUniform}[10^{-3},0.3]$; \\
&max\_depth $\in [3,10]$; subsample $\in [0.5,1.0]$; \\
&max\_features $\in \{\mathrm{sqrt}, \mathrm{log2}, \mathrm{None}\}$ \\
\hline
\multirow{4}{*}{XGB}  &
n\_estimators $\in [50,300]$; \\
&learning\_rate $\sim \mathrm{LogUniform}[10^{-3},0.3]$; \\
&max\_depth $\in [3,10]$; subsample $\in [0.5,1.0]$; \\
&colsample\_bytree $\in [0.5,1.0]$; gamma $\in [0,5]$ \\
\hline
\multirow{4}{*}{LGBM} &
n\_estimators $\in [50,300]$; num\_leaves $\in [20,150]$;\\
&learning\_rate $\sim \mathrm{LogUniform}[10^{-3},0.5]$; \\
& max\_depth $\in [3,15]$; subsample $\in [0.5,1.0]$; \\ 
&colsample\_bytree $\in [0.5,1.0]$ \\
\hline
\end{tabular}
\end{table}

\begin{table}[!h]
\centering
\caption{Architectures of generative models}
\label{tbl:CGAN_CVAE_architecture}
\begin{tabular}{|l|l|}
\hline
\multicolumn{1}{|c|}{\textbf{Model}}                      & \multicolumn{1}{c|}{\textbf{Network (MLP)}}                                                                                                                                                                   \\ \hline
\begin{tabular}[c]{@{}l@{}}CGAN\end{tabular} & \begin{tabular}[c]{@{}l@{}}$G: [z(100), c(20)] \rightarrow 128 \rightarrow 256 \rightarrow 200$\\ $D: [x(200), c(20)] \rightarrow 128 \rightarrow 64 \rightarrow 1$\end{tabular}                              \\ \hline
CVAE                                                      & \begin{tabular}[c]{@{}l@{}}Encoder: $[x(200), c(20)] \rightarrow 256 \rightarrow 128 \rightarrow (\mu,\log\sigma^2)$\\ Decoder: $[z(64), c(20)] \rightarrow 128 \rightarrow 256 \rightarrow 200$\end{tabular} \\ \hline
\end{tabular}
\vspace{-2.5ex}
\end{table}

\begin{table}[!h]
\small
\centering
\caption{Best hyperparameters found by Optuna (500 trials, 5-fold cross-validation). }
\label{tbl:hyperparameters}
\begin{tabular}{|c|c|c|l|}
\hline
\textbf{Region}       & \textbf{AoA} & \textbf{Model} & \textbf{Best hyperparameters}                                                                                                                                                           \\ \hline
\multirow{4}{*}{LoS}  & MUSIC        & RF             & \begin{tabular}[l]{@{}l@{}}``n\_estimators": 297,\\ ``max\_depth": 27,\\ ``max\_features": "log2",\\ ``min\_samples\_split": 3,\\ ``min\_samples\_leaf": 2\end{tabular}                      \\ \cline{2-4} 
                      & ESPRIT       & SVM            & \begin{tabular}[l]{@{}l@{}}``kernel": ``rbf",\\ ``C": 58.5476005441,\\ ``gamma": 0.02981714257685\end{tabular}                                                                      \\ \hline
\multirow{6}{*}{NLoS} & MUSIC        & GBM            & \begin{tabular}[l]{@{}l@{}}``n\_estimators": 256,\\ ``learning\_rate": 0.0585218637465,\\ ``max\_depth": 10,\\ ``subsample": 0.770060357471452,\\ ``max\_features": ``log2"\end{tabular}  \\ \cline{2-4} 
                      & ESPRIT       & GBM            & \begin{tabular}[l]{@{}l@{}}``n\_estimators": 295,\\ ``learning\_rate": 0.0500295940890,\\ ``max\_depth": 10,\\``subsample": 0.8695777940266635,\\ ``max\_features": "log2"\end{tabular} \\ \hline
\end{tabular}
\end{table}

\begin{table}[!h]
\small
\centering
\setlength\tabcolsep{3pt} 
\caption{The detailed performance of six incremental models is evaluated in the LoS condition with MUSIC / ESPRIT, including warm-up training accuracy (Training Acc), online inference accuracy (Online Acc), acceptance rate (Accept. Rate), and forgetting rate.}
\label{tbl:online_result_2_details}
    \begin{subtable}{0.475\textwidth}
        \centering
        \caption{LoS + MUSIC}
        \begin{tabular}{|c|c|c|c|c|}
		\hline
		\textbf{Model} & \textbf{Training Acc} & \textbf{Online Acc} & \textbf{Accept. Rate} & \textbf{Forgetting Rate} \\ \hline
		\textbf{AMF}   & \textbf{0.7559}       & \textbf{0.9432}     & \textbf{0.7286}          & \textbf{0.0343}          \\ \hline
		ARF            & 0.7269                & 0.8230              & 0.7996                   & 0.1231                   \\ \hline
		GNB            & 0.7360                & 0.7790              & 1.0000                   & 0.1455                   \\ \hline
		HAT            & 0.7215                & 0.7775              & 1.0000                   & 0.1433                   \\ \hline
		HT             & 0.7360                & 0.7782              & 1.0000                   & 0.1463                   \\ \hline
		SRP            & 0.7251                & 0.8161              & 0.8962                   & 0.1239                   \\ \hline
		\end{tabular}
    \end{subtable}
    \\
    \begin{subtable}{0.475\textwidth}
        \centering
        \caption{LoS + ESPRIT}
        \begin{tabular}{|c|c|c|c|c|}
		\hline
		\textbf{Model} & \textbf{Training Acc} & \textbf{Online Acc} & \textbf{Accept. Rate} & \textbf{Forgetting Rate} \\ \hline
		\textbf{AMF}   & \textbf{0.7993}       & \textbf{0.9383}     & \textbf{0.7280}          & \textbf{0.0427}          \\ \hline
		ARF            & 0.7143                & 0.8244              & 0.8236                   & 0.1201                   \\ \hline
		GNB            & 0.7360                & 0.7790              & 1.0000                   & 0.1455                   \\ \hline
		HAT            & 0.7215                & 0.7775              & 1.0000                   & 0.1433                   \\ \hline
		HT             & 0.7360                & 0.7782              & 1.0000                   & 0.1463                   \\ \hline
		SRP            & 0.7071                & 0.8423              & 0.8511                   & 0.1108                   \\ \hline
		\end{tabular}
    \end{subtable}
\end{table}

\begin{table}[!h]
\small
\centering
\setlength\tabcolsep{3pt} 
\caption{The detailed performance of six incremental models is evaluated in the NLoS condition with MUSIC / ESPRIT, including warm-up training accuracy (Training Acc), online inference accuracy (Online Acc), acceptance rate (Accept. Rate), and forgetting rate.}
\label{tbl:online_result_2_details1}
    \begin{subtable}{0.475\textwidth}
        \centering
        \caption{NLoS + MUSIC}
        \begin{tabular}{|c|c|c|c|c|}
		\hline
		\textbf{Model} & \textbf{Training Acc} & \textbf{Online Acc} & \textbf{Accept. Rate} & \textbf{Forgetting Rate} \\ \hline
		\textbf{AMF}   & \textbf{0.6559}       & \textbf{0.9409}     & \textbf{0.6290}          & \textbf{0.0248}          \\ \hline
		ARF            & 0.5448                & 0.7130              & 0.5391                   & 0.1838                   \\ \hline
		GNB            & 0.5573                & 0.6135              & 0.9994                   & 0.2114                   \\ \hline
		HAT            & 0.5609                & 0.6040              & 0.9982                   & 0.2134                   \\ \hline
		HT             & 0.5538                & 0.6153              & 0.9994                   & 0.2098                   \\ \hline
		SRP            & 0.5573                & 0.7282              & 0.7729                   & 0.1723                   \\ \hline
		\end{tabular}
    \end{subtable}%
    \\
    \begin{subtable}{0.475\textwidth}
        \centering
        \caption{NLoS + ESPRIT}
        \begin{tabular}{|c|c|c|c|c|}
		\hline
		\textbf{Model} & \textbf{Training Acc} & \textbf{Online Acc} & \textbf{Accept. Rate} & \textbf{Forgetting Rate} \\ \hline
		\textbf{AMF}   & \textbf{0.6631}       & \textbf{0.9412}     & \textbf{0.6020}          & \textbf{0.0258}          \\ \hline
		ARF            & 0.5430                & 0.6962              & 0.5359                   & 0.1874                   \\ \hline
		GNB            & 0.5573                & 0.6135              & 0.9994                   & 0.2114                   \\ \hline
		HAT            & 0.5609                & 0.6040              & 0.9982                   & 0.2134                   \\ \hline
		HT             & 0.5538                & 0.6153              & 0.9994                   & 0.2098                   \\ \hline
		SRP            & 0.5573                & 0.7534              & 0.7539                   & 0.1586                   \\ \hline
		\end{tabular}
    \end{subtable}
\end{table}

\begin{table}[th]
\centering
\caption{Prototypical network architecture specification}
\label{tbl:protonet}
\begin{tabular}{|l|l|}
\hline
\multicolumn{1}{|c|}{\textbf{Component}} & \multicolumn{1}{c|}{\textbf{Details}}                                                                                                     \\ \hline
Input feature                            & $x\in\mathbb{R}^{200}$ (AoA feature vector)                                                                                              \\ \hline
Embedding network $f_\theta$             & \begin{tabular}[c]{@{}l@{}}MLP 200 $\rightarrow$ 128 $\rightarrow$ 64 $\rightarrow$ 32 \\ (BatchNorm + ReLU + Dropout(0.3))\end{tabular} \\ \hline
Prototype $p_c$                          & mean embedding of samples in class $c$                                                                                           \\ \hline
Distance                                 & squared Euclidean distance in $\mathbb{R}^{32}$                                                                                          \\ \hline
Classifier                               & softmax over negative distances                                                                                                          \\ \hline
Optimized parameters                     & \begin{tabular}[c]{@{}l@{}}$\theta$ only (embedding network); \\ prototypes are computed per episode\end{tabular}                        \\ \hline
\end{tabular}
\end{table}


\newpage
\bibliographystyle{IEEEtran}
\bibliography{references}

\begin{IEEEbiography}[{\includegraphics[width=1in,height=1.25in,clip,keepaspectratio]{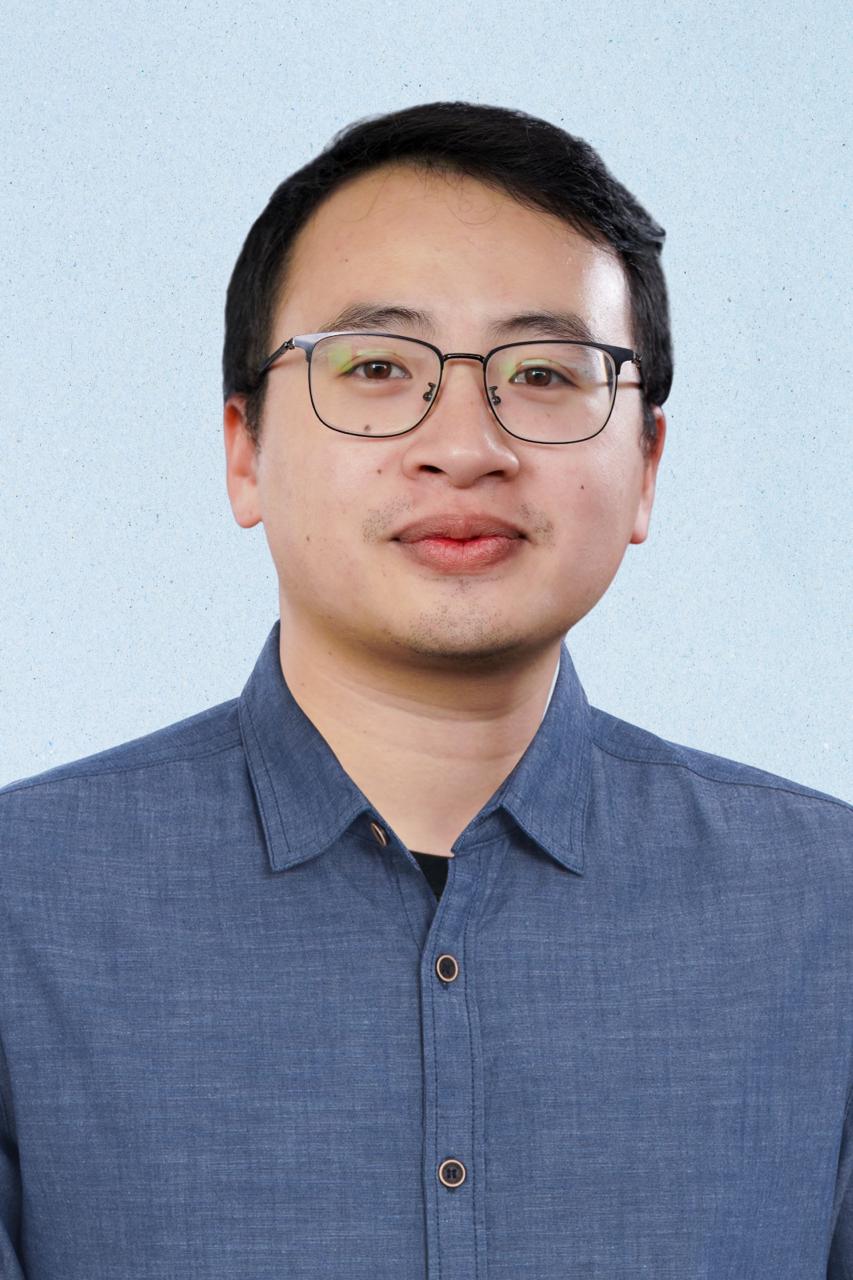}}]
{Bac Trinh-Nguyen}~received his B.Sc. and M.Sc. degrees from the University of Information Technology, VNU-HCM in 2018 and 2021, respectively. He is currently pursuing his Ph.D. within the joint IPAL project, a collaboration between CY Cergy Paris University - specifically the ETIS (Équipe Traitement de l'Information et Systèmes) Laboratory, ICI (Information, Communication, and Images) team - and the Institute for Infocomm Research (I2R), Agency for Science, Technology and Research (A\*STAR), Singapore. His research interests include cybersecurity and the application of machine learning to wireless communications, with a focus on physical layer security (PLS), localization, trustworthiness and privacy.
\end{IEEEbiography}

\begin{IEEEbiography}
[{\includegraphics[width=1in,height=1.25in,clip,keepaspectratio]{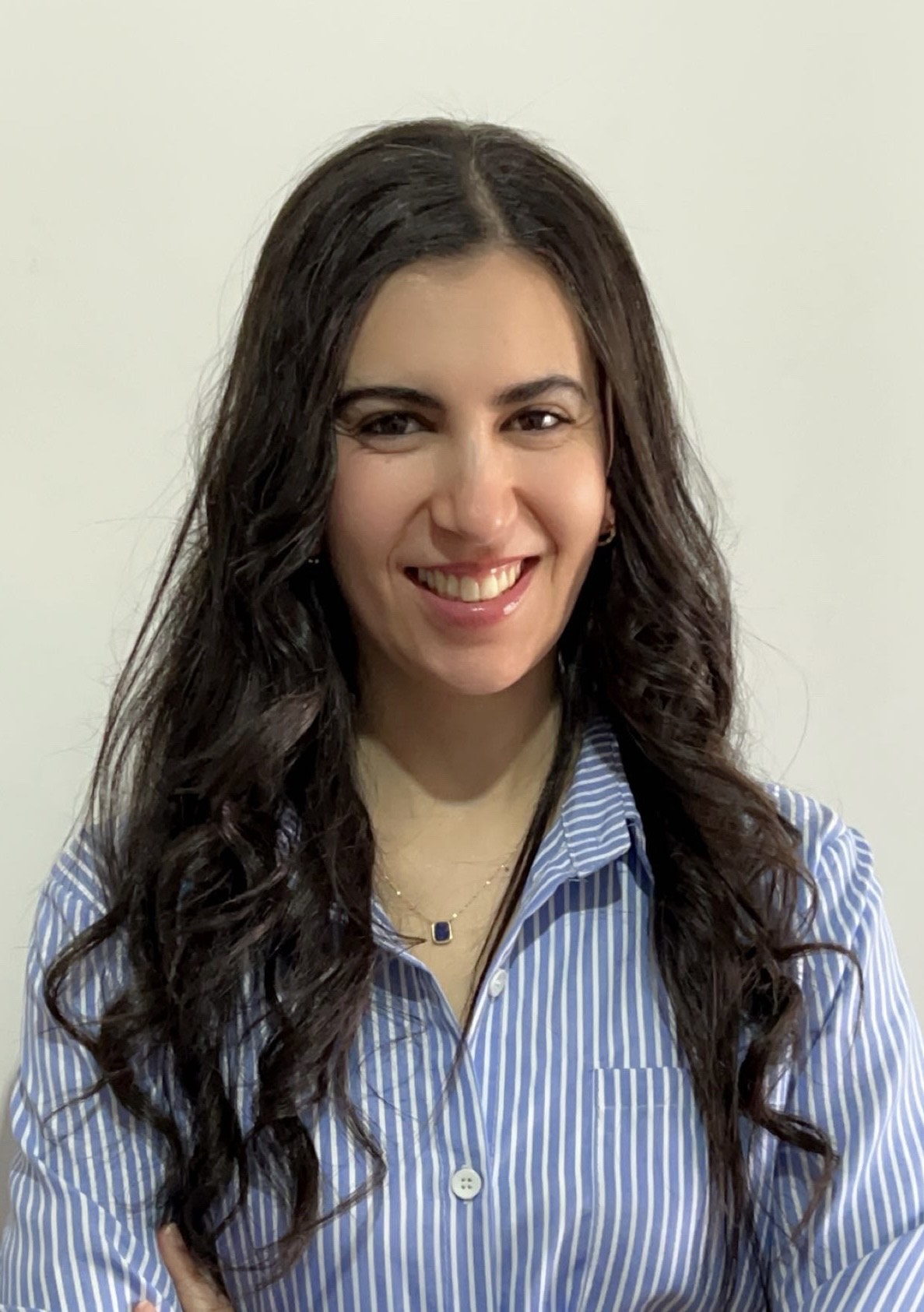}}]
{Sara Berri}~is an Associate Professor (MCF) at the CY Cergy Paris University and a member of the ETIS (Equipe Traitement de l'Information et Systèmes) Laboratory, ICI (Information, Communication, and Images) team, since 2020. Previously, she was a postdoctoral researcher at Télécom Paris (2018-2020), CCN (Cybersecurity for Communication Networks) team of LTCI (Laboratoire Traitement et Communication de l'Information) Laboratory and she defended her PhD thesis in March 2018. She serves as Associate Editor for the IEEE Transactions on Network and Service Management. She is the recipient of the 2025 CY Alliance award: "For women and science". Her research encompasses several aspects of network optimization, O-RAN, localization, resource allocation, privacy in location-based services, intelligent transport systems using optimization, machine learning and game theory.
\end{IEEEbiography}

\begin{IEEEbiography}[{\includegraphics[width=1in,height=1.25in,clip,keepaspectratio]{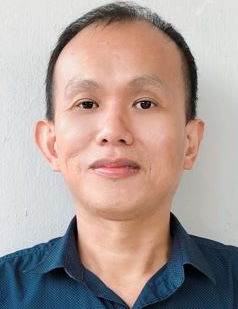}}]{Sin G. Teo}~is a Scientist at Institute for Infocomm Research (I2R),  Agency for Science, Technology and Research (A*STAR), Singapore. Currently, he is a principal investigator of several projects that blends the domains of Artificial Intelligence and Cybersecurity, and leads a team of A.I. for Cybersecurity in the Cybersecurity department.  He obtained the Ph.D. degree from Monash University, Australia in 2016. His research interests include applied cryptography, data privacy and security, malware and network anomaly classification, federated learning, and deep learning.
\end{IEEEbiography}

\begin{IEEEbiography}[{\includegraphics[width=1in,height=1.25in,clip,keepaspectratio]{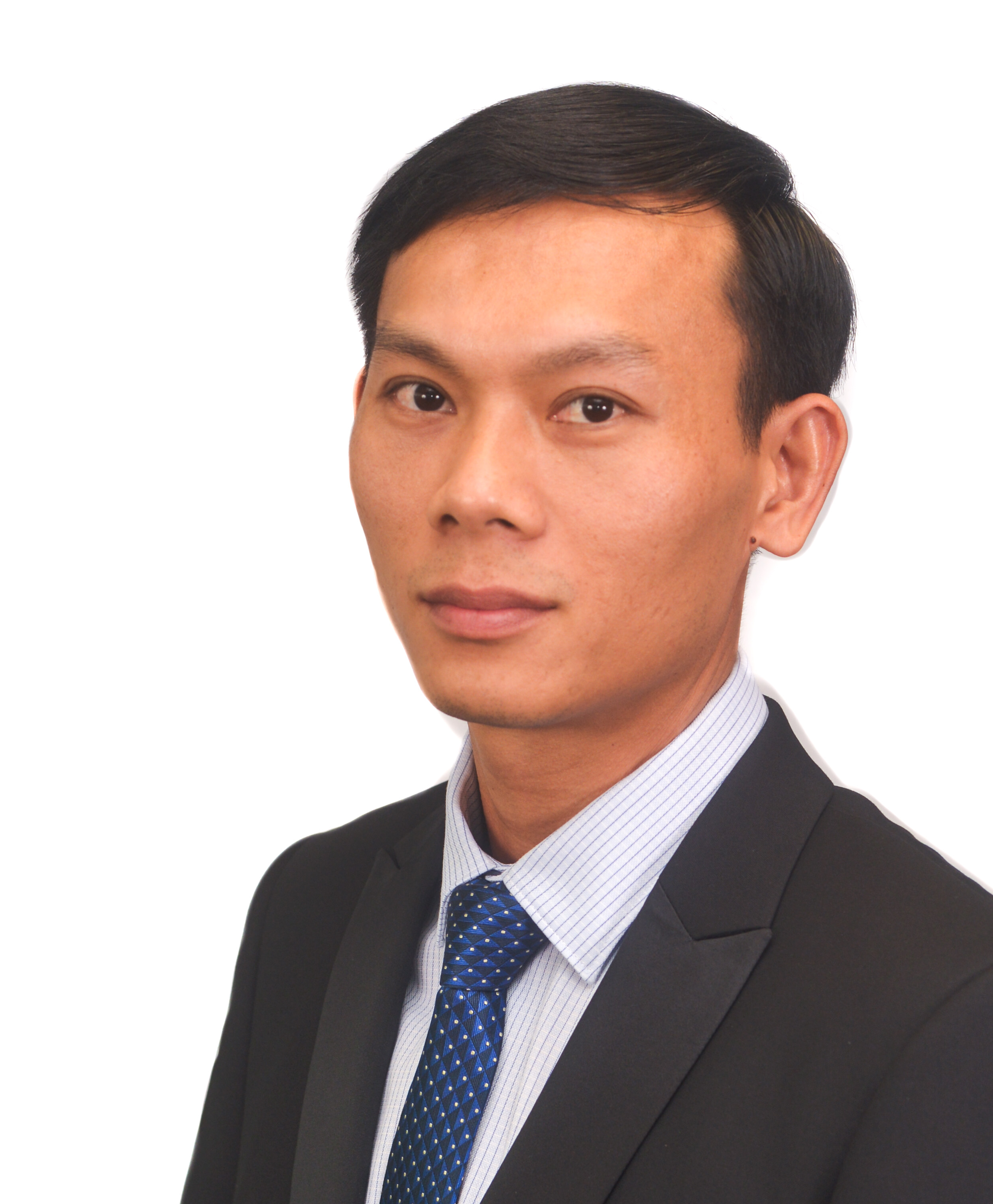}}]{Tram Truong-Huu}~(M'12 -- SM'15) is an Associate Professor at the Singapore Institute of Technology (SIT), Infocomm Technology (ICT) Cluster, currently serving as the Programme Leader for SIT Bachelor of Engineering with Honours in Information and Communications Technology majoring in Information Security. He has been a senior computer scientist at the Institute for Infocomm Research (I2R), the Agency for Science, Technology and Research (A*STAR), since May 2019, and held a joint appointment from August 2021 to  August 2025. He received his Ph.D. degree in computer science from the University of Nice - Sophia Antipolis (now Côte d’Azur University), France, in December 2010. From January 2011 to June 2012, he held a post-doctoral fellowship at the French National Center for Scientific Research (CNRS), France. He worked at the National University of Singapore as a research fellow from July 2012 and then senior research fellow from January 2017. His research interests focus on federated learning and the application of artificial intelligence to cybersecurity and next-generation networking. He has been a member of the IEEE since 2012 and a senior member since 2015.
\end{IEEEbiography}

\begin{IEEEbiography}
[{\includegraphics[width=1in,height=1.25in,clip,keepaspectratio]{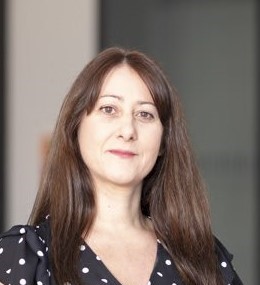}}]
{Arsenia Chorti}~is a Professor at the École Nationale Supérieure de l'Électronique et de ses Applications (ENSEA) at the ETIS Lab UMR 8051, Research Fellow of the Barkhausen Institut gGmbH and a Visiting Scholar at Princeton University. Her research spans the areas of wireless communications and wireless systems security for 5G and 6G, with a particular focus on physical layer security. Current research topics include: context aware security, 5G / 6G, integrated sensing and communications, machine learning for communications, semantic and goal oriented communications. She is a Senior IEEE Member, and has served as IEEE Distinguished Lecturer (2024-2025), Associate Editor in Chief of the IEEE ComSoc Best Readings, Member of the IEEE INGR on Security, Chair of the IEEE Focus Group on Physical Layer Security (2021-24) and member of the IEEE Teaching Awards Committee (2017-19). She is currently member of various ITU Working Groups including on CGDatasets. She has participated in the reduction of the ITU report M.2516-0 on Future technology trends of terrestrial International Mobile Telecommunications systems towards 2030 and beyond (section on trustworthiness). She has served in the IEEE P1940 Standardization Workgroup on ``Standard profiles for ISO 8583 authentication services". She was selected as one of the ``100 Brilliant and Inspiring Women in 6G" for 3 consecutive years between 2024-26.
\end{IEEEbiography}

\end{document}